\documentclass[twoside]{article}
\usepackage[accepted]{aistats2025}
\usepackage[round]{natbib}

\makeatletter
\newcommand{\maketitlenew}{\@maketitle}
\makeatother

\usepackage[plainruled,titlenumbered]{algorithm2e}
\makeatletter
\newcommand{\removelatexerror}{\let\@latex@error\@gobble}
\makeatother
\usepackage{algorithmic}
\usepackage{listings}
\usepackage{chemarr} 
\usepackage{amssymb}
\usepackage{amsmath}
\usepackage{amsfonts}       
\usepackage{mathtools}
\usepackage{nicefrac}       
\usepackage{amsthm}         
\usepackage{dsfont}

\usepackage{empheq}

\allowdisplaybreaks

\theoremstyle{plain}

\theoremstyle{definition}

\theoremstyle{remark}

\usepackage{graphicx}
\graphicspath{ {figures/} }
\usepackage{epsfig}
\usepackage{subcaption}
\usepackage{float}
\usepackage{wrapfig}

\usepackage[utf8]{inputenc} 
\usepackage[T1]{fontenc}    
\usepackage{microtype}      
\usepackage{textcomp}

\usepackage{booktabs}       
\usepackage{makecell}

\usepackage{color}
\usepackage{xcolor}
\usepackage{xkeyval}

\usepackage{tikz}
\usetikzlibrary{bayesnet}
\usetikzlibrary{positioning}
\usetikzlibrary{arrows}
\usetikzlibrary{calc}

\usetikzlibrary{positioning,fit,shapes,backgrounds,circuits.logic.US}

\tikzstyle{server}=[circle, line width=0.5pt, rounded corners=0.1mm, draw=black!100, fill=tud3a!100]
\tikzstyle{vertex}=[circle, line width=1.5pt, draw=tud0d, fill=white]
\tikzstyle{dispatcher} =[and gate US, line width=0.5pt, draw=black!100, fill=tud0c!100]
\tikzstyle{dotbox} = [draw=white, fill=white, rectangle,  inner sep=10pt, inner ysep=20pt]

\tikzset{three_sided/.style={
		draw=none,rectangle, 
		append after command={
			[shorten <= -0.5\pgflinewidth]
			([shift={(-1.5\pgflinewidth,-0.5\pgflinewidth)}]\tikzlastnode.north west)
			edge([shift={( 0.5\pgflinewidth,-0.5\pgflinewidth)}]\tikzlastnode.north east)
			([shift={( 0.5\pgflinewidth,-0.5\pgflinewidth)}]\tikzlastnode.north east)
			edge([shift={( 0.5\pgflinewidth,+0.5\pgflinewidth)}]\tikzlastnode.south east)
			([shift={( 0.5\pgflinewidth,+0.5\pgflinewidth)}]\tikzlastnode.south east)
			edge([shift={(-1.0\pgflinewidth,+0.5\pgflinewidth)}]\tikzlastnode.south west)
		}
	}
}

\usepackage{paralist} 
\usepackage{amstext} 
\usepackage{array} 
\usepackage{hyperref}       

\usepackage{url}            
\usepackage{xspace}
\usepackage{csquotes}
\usepackage{etoolbox}
\usepackage[shortlabels,inline]{enumitem}  
\newlist{inlineitemize}{enumerate*}{1}
\setlist[inlineitemize]{label=(\roman*)}
\usepackage{calc}

\usepackage[capitalize, nameinlink]{cleveref} 

\usepackage{ifthen}
\usepackage{balance}

\usepackage{todonotes}
\newboolean{todo}
\setboolean{todo}{false} 

\newcommand{\remarkInternal}[4]{\ifthenelse{\boolean{todo}}{\todo[inline, color=#2, caption={2do}, #3]{\begin{minipage}{\textwidth-4pt}\emph{Remark #1:}\\#4\end{minipage}}}{}}

\usepackage{blindtext}
\newboolean{blind}
\setboolean{blind}{false} 
\newcommand{\blindparagraph}[2][]{\ifthenelse{\boolean{blind}}{\blindtext[1]}{}}

\usepackage{hyphenat} 
\newcommand{\eg}{e.g.\@\xspace}
\newcommand{\ie}{i.e.\@\xspace}
\newcommand{\wrt}{w.r.t.\@\xspace}
\newcommand{\rhs}{r.h.s.\@\xspace}

\DeclareMathOperator{\diag}{diag}
\DeclareMathOperator{\tr}{tr}

\DeclareMathOperator*{\argmin}{argmin}

\DeclareMathOperator{\1}{\mathds{1}}

\DeclareMathOperator{\E}{\mathsf{E}}

\DeclarePairedDelimiterX{\expectarg}[1]{[}{]}{%
    \ifnum\currentgrouptype=16 \else\begingroup\fi
    \activatebar#1
    \ifnum\currentgrouptype=16 \else\endgroup\fi
}
\newcommand{\innermid}{\nonscript\;\delimsize\vert\nonscript\;}
\newcommand{\activatebar}{%
    \begingroup\lccode`\~=`\|
    \lowercase{\endgroup\let~}\innermid
    \mathcode`|=\string"8000
}

\DeclarePairedDelimiterX{\expectargleft}[1]{[}{.}{%
    \ifnum\currentgrouptype=16 \else\begingroup\fi
    \activatebar#1
    \ifnum\currentgrouptype=16 \else\endgroup\fi
}

\DeclarePairedDelimiterX{\expectargright}[1]{.}{]}{%
    \ifnum\currentgrouptype=16 \else\begingroup\fi
    \activatebar#1
    \ifnum\currentgrouptype=16 \else\endgroup\fi
}

\DeclareMathOperator{\KL}{\mathsf{KL}}

\newcommand{\KLof}{\KL\klarg}
\DeclarePairedDelimiterX{\klarg}[1]{(}{)}{%
    \ifnum\currentgrouptype=16 \else\begingroup\fi
    \activatediv#1
    \ifnum\currentgrouptype=16 \else\endgroup\fi
}
\newcommand{\innerdiv}{\nonscript\;\delimsize\Vert\nonscript\;}
\newcommand{\activatediv}{%
    \begingroup\lccode`\~=`\|
    \lowercase{\endgroup\let~}\innerdiv
    \mathcode`|=\string"8000
}

\DeclarePairedDelimiterX{\klargleft}[1]{()}{.}{%
    \ifnum\currentgrouptype=16 \else\begingroup\fi
    \activatediv#1
    \ifnum\currentgrouptype=16 \else\endgroup\fi
}

\DeclarePairedDelimiterX{\klargright}[1]{.}{)}{%
    \ifnum\currentgrouptype=16 \else\begingroup\fi
    \activatediv#1
    \ifnum\currentgrouptype=16 \else\endgroup\fi
}

\DeclareMathOperator{\NDis}{\mathcal{N}}

\DeclareMathOperator{\PoisDis}{Pois}

\DeclareMathOperator{\Prob}{P}
\DeclareMathOperator{\Qrob}{Q}

\newcommand{\diff}{\mathop{}\!\mathrm{d}}

\usepackage[nolist]{acronym}

\begin{document}

%

%

\begin{acronym}
    \acro{mjp}[MJP]{Markov jump process}
    \acroplural{mjp}[MJPs]{Markov jump processes}
    \acroindefinite{mjp}{an}{a}
    \acro{crn}[CRN]{chemical reaction network}
    \acro{ode}[ODE]{ordinary differential equation}
    \acroindefinite{ode}{an}{an}
    \acro{sde}[SDE]{stochastic differential equation}
    \acroindefinite{sde}{an}{a}
    \acro{kl}[KL]{Kullback-Leibler}
    \acro{em}[EM]{expectation maximization}
    \acroindefinite{em}{an}{an}
    \acro{estep}[E-step]{expectation step}
    \acroindefinite{estep}{an}{an}
    \acro{mstep}[M-step]{maximziation step}
    \acroindefinite{mstep}{an}{a}
     \acro{mcmc}[MCMC]{Markov chain Monte Carlo}
    \acroindefinite{mcmc}{an}{a}
    \acro{smc}[SMC]{sequential Monte Carlo}
    \acroindefinite{smc}{an}{a}
    \acro{vi}[VI]{variational inference}
    \acro{ads}[ADS]{assumed density smoother}
    \acro{ep}[EP]{expectation propagation}
    \acro{ffbs}[FFBS]{forward-filtering backward-smoothing}
    \acro{ads}[ADS]{assumed density smoother}
\end{acronym}
\twocolumn[

\aistatstitle{Entropic Matching for Expectation Propagation of Markov Jump Processes}

\aistatsauthor{ Yannick Eich \And Bastian Alt \And  Heinz Koeppl }

\aistatsaddress{Department of Electrical Engineering and Information Technology\\\
  Technische Universität Darmstadt
  \\
  \texttt{\{yannick.eich,  heinz.koeppl\}@tu-darmstadt.de} }

  
   ]

\begin{abstract}
We propose a novel, tractable latent state inference scheme for Markov jump processes, for which exact inference is often intractable. Our approach is based on an entropic matching framework that can be embedded into the well-known expectation propagation algorithm.
We demonstrate the effectiveness of our method by providing closed-form results for a simple family of approximate distributions and apply it to the general class of chemical reaction networks, which are a crucial tool for modeling in systems biology.
Moreover, we derive closed-form expressions for point estimation of the underlying parameters using an approximate expectation maximization procedure. 
We evaluate our method across various chemical reaction networks and compare it to multiple baseline approaches, demonstrating superior performance in approximating the mean of the posterior process. Finally, we discuss the limitations of our method and potential avenues for future improvement, highlighting its promising direction for addressing complex continuous-time Bayesian inference problems.
\end{abstract}
\section{INTRODUCTION}
\Acp{mjp} play a crucial role for modeling diverse phenomena in various domains, including finance \citep{mamon2007hidden}, engineering  \citep{bolch2006queueing}, and biology \citep{anderson2015stochastic}. 
In the field of systems biology, \acp{mjp} find particular significance, offering powerful modeling capabilities for complex systems, such as \acp{crn} \citep{anderson2015stochastic}.
By incorporating prior knowledge of the dynamic nature of underlying processes, \ac{mjp} models can efficiently extract valuable insights and facilitate understanding and control of these intricate systems.
This becomes particularly crucial when dealing with latent processes, where only partial information about the desired quantities of interest is available.
The resulting inverse problem presents significant challenges, requiring the solution of the underlying Bayesian filtering and smoothing problem.

Traditional approaches to latent state inference in continuous-time stochastic processes often rely on system approximations using \acp{ode} or \acp{sde}, see, \eg, \citep{gardiner1985handbook}.
However, inference methods based on Kalman filtering and RTS smoothing that exploit linearization procedures, such as the linear noise approximation \citep{komorowski2009bayesian}, can suffer from the limitations of the underlying non-linear rate function.
Similarly, approaches based on the non-linear chemical Langevin equation \citep{gillespie2000chemical} may yield inaccurate results, especially in scenarios characterized by low counting numbers.

In contrast, modeling the process directly using \iac{mjp} model captures discrete state transitions and provides a more accurate representation of the underlying dynamics.
However, latent state inference in \acp{mjp} often requires computationally demanding sampling-based techniques, such as \ac{smc} \citep{doucet2001sequential,golightly2011bayesian}, which suffers from the well known particle degeneracy problem, especially for long trajectories.
More recently, particle \ac{mcmc} methods \citep{andrieu2010particle,golightly2015delayed, lowe2023accelerating} have shown promising results in Bayesian parameter estimation; however, the underlying latent state estimation still relies on \ac{smc}, inheriting its limitations.

Alternatively, deterministic methods based on \ac{vi} provide valuable tools. 
Here, the posterior process is approximated using tractable approximations, such as mean-field \ac{vi} \citep{opper2007variational} or moment-based \ac{vi} \citep{wildner2019moment}. More recently, \citet{seifner2023neural} proposed a neural variational inference method to jointly learn the parameters and the latent state. However, their latent state estimation relies on the integration of the chemical master equation, which is not scalable to large models.

We take a different path by employing a message passing method in continuous-time.
Instead of approximating the distribution on a path-wise level, as seen in variational Bayesian methods, our approach involves approximating the exact message passing scheme and thereby optimizing the posterior marginal distributions.
This allows us to accommodate for the \ac{mjp} dynamics and to embed our method into the \ac{ep} algorithm \citep{minka2001family}.  An implementation of our proposed method is publicly available\footnote{\url{https://github.com/yannickeich/ep4crns}}.

\section{BACKGROUND}
\paragraph{Markov Jump Processes.}
A \acf{mjp} \citep{ethier2009markov} is a continuous-time Markov process $\{X(t) \in \mathcal X \mid t \in \mathbb R_{\geq 0}\}$ on a discrete state space $\mathcal X$.
The Markov property implies that for all $t'>t$ we have
$
    \Prob(X(t') = x(t') \mid \{X(s)=x(s) \mid s \in [0,t]\})=\Prob(X(t') = x(t') \mid X(t)=x(t)).
$
Hence, \iac{mjp} is fully described by an initial distribution  $p_0(x)=\Prob(X(0)=x)$, $\forall x \in \mathcal X$, and its rate function 
\begin{equation*}
  \Lambda(x,x',t) \coloneqq  \lim_{h \to 0} \frac{\Prob(X(t+h)=x'\mid X(t)=x)}{h},
\end{equation*}
for all time points $t \in \mathbb R_{\geq 0}$ and states $x \in \mathcal X \neq x' \in \mathcal X$.
Given these quantities, one can derive a system of \acp{ode} of the time-marginal probability function $p_t(x) \coloneqq \Prob(X(t)=x)$ of the \ac{mjp}, which is given by the differential forward Chapman-Kolmogorov equation for $p_t(x)$, the \emph{master equation},
\begin{equation}
\label{eq:master_eq_prior}
\begin{aligned}
       &\frac{\diff}{\diff t} p_t(x)=[\mathcal L_t p_t](x), && \forall x \in \mathcal X,
\end{aligned}
\end{equation}
with initial distribution $p_0(x)$ at time point $t=0$, where the operator $\mathcal L_t$ is given as
\begin{equation*}
    [\mathcal L_t \phi](x) \coloneqq \sum_{x' \neq x} \Lambda(x',x,t) \phi(x') - \Lambda(x,x',t) \phi(x),
\end{equation*}
for an arbitrary test function $\phi$.
Additionally, an evolution equation for some arbitrary moment functions can be found, by multiplying \cref{eq:master_eq_prior} from the left with a moment function $s(x)$ and summing over all elements $x \in \mathcal X$, this yields
$
     \nicefrac{\diff}{\diff t} \E[s(X(t))]=\E[\mathcal L_t^\dagger s(X(t))],
$
where $\mathcal L_t^\dagger$ is the adjoint operator of $\mathcal L_t$, \wrt the inner product $\langle \phi, \psi \rangle= \sum_{x \in \mathcal X} \phi(x)\psi(x)$ given as
\begin{equation*}
     [\mathcal L_t^\dagger \psi](x) \coloneqq \sum_{x' \neq x} \Lambda(x,x',t) [\psi(x') -\psi(x)],
\end{equation*}
for an arbitrary test function $\psi$. For more on \acp{mjp} see \citep{ethier2009markov, gardiner1985handbook, norris1998markov, del2017stochastic}.

\paragraph{Chemical Reaction Networks.}
We focus on \acfp{crn} \citep{anderson2015stochastic} with the species $\{\mathsf{X}_i \mid i=1, \dots, n \}$ and $k$ reactions given as
\begin{equation*}
\begin{aligned}
 &\underline{\nu}_{1j} \mathsf{X}_1 + \dots +\underline{\nu}_{nj} \mathsf{X}_n \xrightarrow{c_j} \bar{\nu}_{1j}\mathsf{X}_1 + \dots + \bar{\nu}_{nj}\mathsf{X}_n, 
\end{aligned}
\end{equation*}
$\forall j=1,\dots, k$, where $c_j \in \mathbb R_{\geq  0}$ is the reaction rate of the $j$th reaction.
The substrate and product stoichiometries are given by the matrices $\underline{\nu} \in \mathbb N_{0}^{n \times k}$ and $\bar{\nu} \in \mathbb N_{0}^{n \times k}$, respectively. 
The state $X(t) \in \mathcal X$ of the network at time point $t$ is described by the amount of each of the $n$ species $X(t)=[X_1(t),\dots, X_n(t)]^\top$, with $X_i(t) \in \mathcal X_i \subseteq \mathbb N_{0}$, $\forall i=1,\dots, n$, and $\mathcal X = \bigtimes_{i=1}^n \mathcal X_i\subseteq\mathbb N_0^n$. 
For \acp{crn} one assumes that the stochastic process $\{X(t) \mid t \in \mathbb R_{\geq 0}\}$ is \iac{mjp}, with rate function
\begin{equation}
\label{eq:crn_rate_fun}
    \Lambda(x,x', t) = \sum_{j=1}^k \1(x'=x + \nu_j)\lambda_j(x),
\end{equation}
where the change vector $\nu_j \in \mathbb Z^n$ corresponding to the $j$th reaction is 
$
    \nu_j = \bar{\nu}_{\cdot j}-\underline{\nu}_{\cdot j}
$.
We assume that the propensity function $\lambda_j(x)$ corresponding to the $j$th reaction is given by mass action kinetics as
\begin{equation}
\label{eq:mass_action_hazard}
    \lambda_j(x) = \bar c_j \prod_{i=1}^{n}  \binom{x_i}{\underline{\nu}_{ij}} = c_j \prod_{i=1}^{n} (x_i)_{\underline{\nu}_{ij}},
\end{equation}
where the factors $\{\underline{\nu}_{1j}!,\dots, \underline{\nu}_{nj}!\}$ are absorbed in the $j$th reaction rate coefficient as 
$\
   c_j= \nicefrac{\bar c_j}{\prod_{i=1}^n \underline{\nu}_{ij}!}
$ and $(m)_n \coloneqq \frac{m!}{(m-n)!}= \prod_{k=0}^{m-1} (n-k)$ denotes the falling factorial. For more on \acp{crn} see~\citep{gardiner1985handbook, anderson2015stochastic, wilkinson2018stochastic}.

\section{EXACT INFERENCE FOR LATENT MARKOV JUMP PROCESSES}
We consider continuous-discrete inference \citep{maybeck1982stochastic,huang2016reconstructing,sarkka2019applied}  for latent \acp{mjp}.
The model is given by the latent \ac{mjp} $\{X(t) \in \mathcal X \mid t \in \mathbb R_{\geq 0}\}$ on $\mathcal X$ characterized via its rate function $\Lambda(x,x',t)$ and its initial probability distribution $p_0(x)$.
The latent state $X(t)$ is not directly observed, rather we consider a discrete-time observation model with $N$ observations $\{Y_1,\dots, Y_N\}$ at time points $t_1< t_2<t_3< \dots<t_N$ as
\begin{equation*}
\begin{aligned}
    &Y_i \mid \{X(t_i)=x\} \sim p(y_i\mid x), &&\forall i=1,\dots, N.
    \end{aligned}
\end{equation*}

The problem of inferring the latent \ac{mjp} $X(t)$ at time point $t$ given some observations $y_{[0,T]} \coloneqq \{y_1,\dots, y_N\}$ in a time interval $[0,T]$ is cast as a continuous-time Bayesian filtering and smoothing problem \citep{pardoux1981non,anderson1983smoothing,sarkka2019applied}, similar to the discrete-time setting \citep{sarkka2013bayesian}. 
To this end, the two elementary objects are the filtering $\pi_t(x)$ and smoothing distribution $\tilde \pi_t(x)$, which are defined as
\begin{equation*}
\begin{aligned}
    &\pi_t(x)\coloneqq \Prob(X(t)=x \mid y_{[0,t]}), \\
    &\tilde \pi_t(x) \coloneqq \Prob(X(t)=x \mid y_{[0,T]}), &&\forall x \in \mathcal X.
\end{aligned}
\end{equation*}
For the filtering distribution  $\pi_t(x)$ we condition on the set $y_{[0,t]} \coloneqq \{y_1,\dots, y_K\}$ of $K=\sum_{i=1}^N \1(t_i \leq t)$ observations in the interval $[0,t]$ up until time point $t$ and for the smoothing distribution $\tilde \pi_t(x)$ we consider all observations $y_{[0,T]}$ in the whole interval $[0,T]$ up until time point $T$.
The filtering distribution can be computed recursively. 
First, it is easy to notice that the filtering distribution at the initial time point $t=0$ is the initial distribution of the latent \ac{mjp}, \ie, $\pi_0(x)=\Prob(X(0)=x)=p_0(x)$.
Additionally, a standard result in continuous-discrete filtering is that the filtering distribution in between observations follows the latent prior dynamics and at the observation time points is subject to discrete-time updates \citep{huang2016reconstructing,sarkka2019applied}.
Hence, we have a system of \acp{ode} with reset conditions as
\begin{equation}
\label{eq:filter_exact}
\begin{aligned}
        &\pi_0(x)=p_0(x),\\
        &\frac{\diff}{\diff t} \pi_t(x) = [\mathcal L_t \pi_t](x),\\
        &\pi_{t_i}(x)=\frac{p(y_i\mid x) \pi_{t_i^-}(x)}{\sum_{x'\in \mathcal X} p(y_i \mid x') \pi_{t_i^-}(x')},
\end{aligned}
\end{equation}
 $\forall x \in \mathcal X$ and $i = 1,\dots,N$, where throughout we denote by $\phi(t_i^-)\coloneqq \lim_{t \nearrow t_i} \phi(t)$ the limit from the left for a left-continuous function $\phi$.
Note, that the filtering distribution $\pi_t(x)$ is a càdlàg process in time $t$. 
There are multiple options to compute an evolution equation for the smoothing distribution. 
Often a backward-filtering and subsequent forward-smoothing approach is used for continuous-time systems, see, \eg, \citep{archambeau2011approximate,wildner2019moment,mider2021continuous}.
For completeness, we discuss this approach in \cref{app:derivation_forward_smoothing}.
However, later on we exploit a different approach, which considers a \ac{ffbs} scheme. 
First, it is easy to notice that the smoothing distribution has an end point condition, as the smoothing distribution at time point $t=T$ is equal to the filtering distribution, \ie, $\tilde \pi_T(x)= \Prob(X(T)=x \mid y_{[0,T]})=\pi_T(x)$.
For the \ac{ffbs} scheme it is shown in \citep{anderson1983smoothing} that the smoothing distribution follows an \ac{ode} with the filter end-point condition as
 \begin{equation}
 \label{eq:smoother_exact}
 \begin{aligned}
      &\tilde \pi_T(x)=\pi_T(x), &&\frac{\diff}{\diff t} \tilde \pi_t(x) = -[\tilde{\mathcal L}_t \tilde \pi_t](x),
 \end{aligned}
 \end{equation}
 where the backward smoothing operator $\tilde{\mathcal L}_t$ and the corresponding backward smoothing rate function $\tilde \Lambda(x',x,t)$ is given as
 \begin{equation*}
 \begin{aligned}
      &[\tilde{\mathcal L}_t \phi](x) \coloneqq \sum_{x' \neq x} \tilde \Lambda(x',x,t) \phi(x') - \tilde \Lambda(x,x',t) \phi(x),\\
      & \tilde \Lambda(x,x',t)= \Lambda(x',x,t) \frac{\pi_t(x')}{\pi_t(x)},
 \end{aligned}
 \end{equation*}
where $\phi$ is an arbitrary test function and $\tilde \Lambda(x,x',t)$ is defined for states $x \neq x'$ with $ x,x' \in \mathcal X$.
The backward smoothing rate function $\tilde \Lambda(x',x,t)$ depends on the filtering distribution $\pi_t(x)$.
The ratio of filtering distributions appearing in the backward smoothing rate can be seen as a correction factor, when computing the dynamics of a backward Markov process \citep{elliott1986reverse,van2007filtering}, similar to the score correction appearing for continuous state space systems \citep{anderson1982reverse}.
 Therefore, by integrating the filtering distribution forward in time and computing the smoothing distribution backward in time, we can solve the filtering and smoothing problem. 
 However, the substantial down-side of the exact filtering and smoothing equations is that they are intractable. 
 This can easily be seen, by noticing that the distributions are defined on the state space $\mathcal X$. 
 Hence, all sums go over $\vert \mathcal X \vert$ elements and we have to solve the $\vert \mathcal X \vert$-dimensional system of \acp{ode} \labelcref{eq:filter_exact} forwards in time and backwards in time in \cref{eq:smoother_exact}.
 For example, for \acp{crn} with a state space $\mathcal X= \mathbb N_0^n$ this is even infinite dimensional. 
 Additionally, even when truncating the state space, the complexity still scales exponentially in the dimensionality $n$.
 Hence, we are required to use an approximate inference method. \looseness -1
 \section{APPROXIMATE INFERENCE FOR LATENT MARKOV JUMP PROCESSES}
\begin{figure}
    \centering
    \includegraphics[width=.48\textwidth]{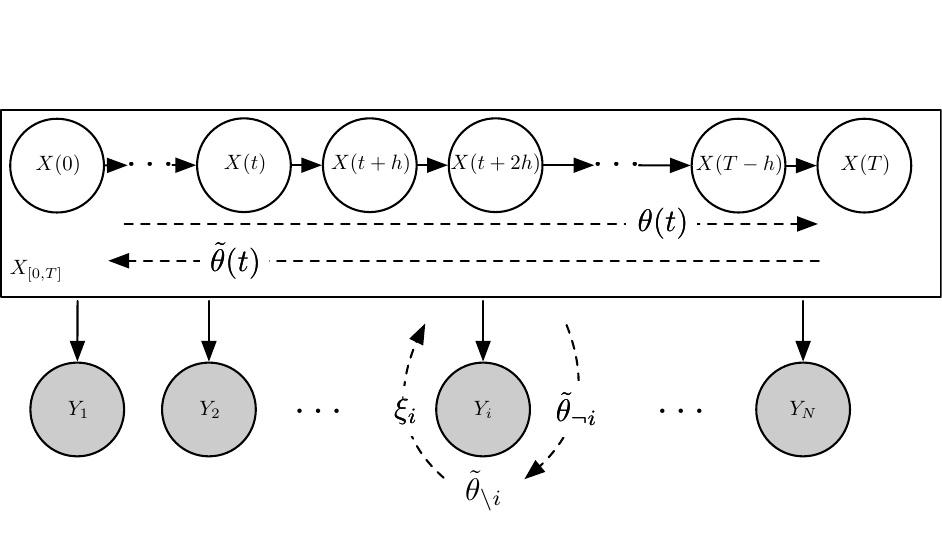}
    \caption{Probabilistic graphical model and approximate inference scheme for \acp{mjp}.
    The continuous-time Markov process $X_{[0,T]}=\{X(t) \mid t \in [0,T]\}$ emits the observations $\{Y_1,\dots, Y_N\}$.
    The approximate inference scheme consists of a continuous-time message passing algorithm, depicted by the dashed lines.
    }
    \label{fig:pgm}
\end{figure}

 We use an approximate inference method to perform latent state inference for \ac{mjp} models.
 To this end we approximate both the filtering distribution and the smoothing distribution in a \ac{ffbs} scheme by using a parametric distribution $q(x\mid \theta)$ as $\pi_t(x) \approx q(x\mid \theta(t))$ and $\tilde \pi_t(x) \approx q(x\mid \tilde \theta(t))$, where $\theta(t) \in \Theta \subseteq \mathbb R^p$ and $\tilde \theta(t)\in \Theta $ are variational parameters for the filtering and smoothing distribution, respectively.
To find those variational parameters we use an assumed density method known as entropic matching \citep{PhysRevE.87.022719,bronstein2018variational}.
Therefore, we derive approximate filtering and smoothing equations for latent \acp{mjp}. 
Further, we exploit an \ac{ep} algorithm \citep{minka2001family} for inference and, therefore, extend the method of \citet{cseke2016expectation} from diffusion processes to \acp{mjp}.
A probabilistic graphical model and the approximate inference scheme is depicted in \cref{fig:pgm}

\subsection{Entropic Matching for Filtering and Smoothing}
We find the variational filtering parameters $\theta(t)$ by considering the minimization of the inclusive \ac{kl} divergence as
$
    \theta(t+h)=\argmin_{\theta'} \KLof{\pi_{t+h}(x)|q(x\mid \theta')}
$,
where $\pi_{t+h}(x)$ is the propagated filtering distribution and $h$ is a small time step. 
By considering that the previous filtering distribution at time point $t$ is approximately $q(x\mid \theta(t))$ and taking the continuous-time limit $h \to 0$ this can be shown \citep{bronstein2018variational} to converge to the \ac{ode} in parameter space as
 \begin{equation}
 \label{eq:pred_step_approx_filter}
     \frac{\diff}{\diff t} \theta(t)= F(\theta(t))^{-1} \E_{q(x \mid {\theta}(t))}\left[\mathcal L_t^\dagger \nabla_\theta \log q(X \mid \theta(t)) \right],
 \end{equation}
 where $F(\theta)\coloneqq - \E_{q(x \mid \theta)}\left[\nabla_\theta \nabla_\theta^\top \log q(X \mid \theta)\right]$ is the Fisher information matrix of the parametric distribution $q(x \mid \theta)$.
At the observation time points, we have discrete-time resets as
 \begin{equation}
 \begin{aligned}
  \label{eq:update_step_approx_filter}
      \theta(t_i)= \argmin_{\theta'} \KLof*{ p(y_i\mid x) q(x\mid \theta(t_i^-))| q(x\mid \theta')},
 \end{aligned}
 \end{equation}
 $\forall i = 1,\dots, N$, and the initial condition can be computed as 
 \begin{equation}
     \label{eq:initial_condition_approx_filter}
     \theta(0)= \argmin_{\theta'} \KLof*{ p_0(x)| q(x\mid \theta')}.
 \end{equation}
The derivation is given for completeness in \cref{app:derivation_filter}.
%
 Analogously, we use the entropic matching method backwards in time for the smoother $ \tilde \pi_t(x) \approx q(x \mid \tilde{\theta}(t))$.
 Here, we compute for a small time step $h$ the time-backwards update
 $
     \tilde \theta(t-h)=\argmin_{\tilde \theta} \KLof{\tilde \pi_{t-h}(x) | q(x \mid \tilde \theta)}
 $.
Considering again that $\tilde \pi_t(x) \approx q(x \mid \tilde \theta(t))$ and carrying out the continuous-time limit $h \to 0$ we derive the corresponding equation for the smoother as
 \begin{equation}
 \label{eq:smoothing_step_approx_filter}
     \frac{\diff}{\diff t} \tilde \theta(t)= - F(\tilde \theta(t))^{-1} \E_{q(x \mid \tilde{\theta}(t))}\left[\tilde{\mathcal L}_t^\dagger \nabla_\theta \log q(X \mid \tilde \theta(t)) \right],
 \end{equation}
 with end-point condition $\tilde \theta(T)=\theta(T)$.
 The full derivation is given in \cref{app:derivation_smoothing}.

 \subsection{Expectation Propagation for Latent Markov Jump Processes}

To further improve our approximation of the posterior, we can use the standard \ac{ep} algorithm \citep{minka2001family, bishop2006pattern, wainwright2008graphical,cseke2016expectation}.
We give here only a brief explanation of the algorithm, for a more detailed explanation we refer the reader to the provided resources.
In the previously mentioned \ac{ffbs} scheme, the contribution of the $i$th measurement to the posterior is included in the $i$th update of the filter as
\begin{equation}
\begin{aligned}
    &\theta(t_i)=\theta(t_i^-)+\xi_i, &&\forall i=1,\dots, N,
\end{aligned}
\label{eq:update_step_approx_filter_sites}
\end{equation}
with the likelihood contribution
\begin{equation*}
\begin{aligned}
    \xi_i = \Bigg(& \argmin_{\theta'} \KLof*{ p(y_i\mid x) q(x\mid \theta(t_i^-)) \,\Big\|\, q(x\mid \theta')} \Bigg)\\
     &- \theta(t_i^-).
\end{aligned}
\end{equation*}

The \ac{ep} algorithm optimizes the likelihood contributions $\xi_i$, also called site parameters, in an iterative manner. 
In each iteration $j$, the first step to update the site parameters is to calculate the \emph{cavity parameters} $\tilde \theta_{\neg i}^{(j)}$, $\forall i=1,\dots, N$, by excluding the $i$th likelihood contribution from the current approximate posterior estimate as
 \begin{equation}
 \begin{aligned}
      &\tilde \theta_{\neg i}^{(j)}={\tilde \theta(t_i^-)}^{(j)}-\xi_i^{(j)}, &&\forall i=1,\dots, N.
 \end{aligned}
 \label{eq:cavity_mjp_update}
 \end{equation}
 Next, the observation factors are incorporated to yield new approximate posterior parameters $\tilde \theta_{\setminus i}^{(j)}$ for the $i$th factor by computing the \emph{tilted distribution} and projecting it back to the parametric family as
 \begin{equation}
 \begin{aligned}
     &\tilde \theta_{\setminus i}^{(j)}=\argmin_{\theta'} \KLof*{ p(y_i\mid x) q(x\mid \tilde \theta_{\neg i}^{(j)})| q(x\mid \theta')}, 
 \end{aligned}
 \label{eq:tilted_mjp_update}
 \end{equation}
$\forall i=1,\dots, N.$ Subsequently, a revised site parameter is found such that, when combined with the cavity parameters, it yields the new approximate posterior. This is done by the subtraction of the cavity parameter as
 \begin{equation}
 \begin{aligned}
     &\tilde{\xi}_i^{(j)}= \tilde \theta_{\setminus i}^{(j)} - \tilde \theta_{\neg i}^{(j)}, &&\forall i=1,\dots, N.
 \end{aligned}
  \label{eq:site_mjp_update}
\end{equation}
Finally, to have a convergent algorithm, a damped message passing strategy, with learning rate parameter $0<\epsilon \leq 1$, is performed as 
\begin{equation}
\begin{aligned}
    &\xi_i^{(j+1)}=(1-\epsilon)\xi_i^{(j)} +\epsilon  \tilde \xi_i^{(j)}, &&\forall i=1,\dots, N.
\end{aligned}
 \label{eq:damped_site_mjp_update}
\end{equation}
The algorithm is then iterated by computing the posterior estimates ${\tilde \theta(t_i^-)}^{(j)}$, $\forall i=1,\dots, N$, and subsequently the new site parameters $\xi_i^{(j+1)}$, $\forall i=1,\dots, N$, for iteration steps $j=1,2,\dots$ until convergence.
For initialization, we set the site parameters to zero, \ie, $\xi_1^{(1)}=0,\dots, \xi_N^{(1)}=0$, which corresponds to setting the initial approximate posterior to an approximate prior distribution.
The \ac{ep} algorithm can be shown to be a sensible algorithm, in the sense that it maximizes a tractable approximation of the intractable log marginal likelihood $\log p(y_{[0,T]})$ \citep{cseke2016expectation} in the form of a fixed point equation for a relaxed variational principle of the intractable problem; for more information, see \citep{wainwright2008graphical}.

\subsection{Parameter Learning}
While our primary focus is on latent state inference, we utilize our approximate inference scheme to develop an approximate \ac{em} algorithm for parameter estimation, similar to the discrete-time case \citep{heskes2003approximate}.
Therefore, we estimate the parameters $\phi$ of the system by maximizing a lower bound  $\mathrm L$ of the marginal likelihood as
\begin{equation*}
\begin{aligned}
    \log p(y_{[0,T]} \mid \phi ) &\geq \mathrm L(\phi) \\
    &\coloneqq \E_{\Qrob}\left[ \log \frac{\diff\Prob}{\diff \Qrob}(X_{[0,T]}, y_{[0,T]}, \phi) \right],
    \end{aligned}
\end{equation*}
where $\frac{\diff \Prob}{\diff \Qrob}$ is the Radon-Nikodym derivative of the path measure $\Prob(X_{[0,T]}\in \cdot, Y_{[0,T]} \in \cdot \mid \phi)$ for the latent paths $X_{[0,T]} \coloneqq \{X(t) \mid t \in [0,T]\}$ and all observations $Y_{[0,T]}$ \wrt an approximating probability measure $\Qrob(X_{[0,T]} \in \cdot)$ over latent paths and the Lebesgue measure over all observations $Y_{[0,T]}$.
Note, that compared to the discrete-time case, we have to express this bound in terms of the Radon-Nikodym derivative, compared to an expression using probability densities \wrt the Lebesgue measure, since we have an uncountable number of random variables $\{X(t) \mid t \in [0,T]\}$ for which there is no Lebesgue measure, see, \eg, \citep{matthews2016sparse}.
A standard result, see, \eg, \citep{bishop2006pattern}, is that the bound can be iteratively maximized by computing \iac{estep} and \iac{mstep}.
To compute the bound one has to exploit Girsanov's theorem \citep{kipnis1998scaling, hanson2007applied}, but also other derivations can be found in the literature, \eg, \citep{opper2007variational,cohn2010mean}.
We compute this bound in \cref{app:param_learning_mjp} in terms of the smoothing and filtering distribution, $\tilde \pi_t(x)$ and $\pi_t(x)$, respectively.
Hence, by replacing these with the approximate smoothing and filtering distribution, $q(x \mid \tilde \theta(t))$ and $q(x \mid \theta(t))$, respectively, we can compute the \ac{mstep} of the algorithm.
This is achieved by maximizing the bound $\mathrm L$ which computes to 
\begin{equation}
\label{eq:marginal_ll_bound}
    \begin{multlined}
        \mathrm L(\phi)= \E_{q(x\mid \tilde\theta(0))}\left[\log \frac{p_0(X \mid \phi)}{q(X\mid \tilde \theta(0))} \right] 
        + \int_{0}^T \E_{q(x \mid \tilde \theta(t))}\bigg[\\
        \sum_{x' \neq X} \left\{\Lambda(X,x',t) \frac{q(x' \mid \tilde \theta(t))}{q(X \mid \tilde \theta(t))}   \frac{q(X\mid  \theta(t))}{q(x' \mid \theta(t))} \right.\\
        \cdot  \log \frac{\Lambda(X,x',t \mid \phi)}{\Lambda(X,x',t)}
        - \Lambda(X,x',t\mid \phi)\bigg\} \bigg] \diff t \\
        + \sum_{i=1}^N \E_{q(x \mid \tilde \theta(t_i))}\left[\log p(y_i \mid X, \phi)\right] + C, \qquad \qquad
    \end{multlined}
\end{equation}
where $\Lambda(x,x',t)$ are the rates of the approximating distribution $\Qrob$ independent of $\phi$, $\Lambda(x,x',t\mid \phi)$ are the parameter dependent rates of the measure $\Prob$ and $C$ is parameter independent constant.
Therefore, the approximate \ac{em} algorithm can be achieved by carrying out the \ac{estep}, which consists of computing the approximate distributions $q(x \mid \theta(t))$ and $q(x \mid \tilde \theta(t))$ for all $t\in [0,T]$, keeping the parameters $\phi$ fixed.
The \ac{mstep} then consists of optimizing \cref{eq:marginal_ll_bound} \wrt the parameters $\phi$, keeping the approximating distributions $q(x \mid \theta(t))$ and $q(x \mid \tilde \theta(t))$ fixed.
For more details, see \cref{app:param_learning_mjp}.

\section{APPLICATION TO LATENT CHEMICAL REACTION NETWORKS}
In this section, we apply the approximate inference algorithm to \acp{crn}. 
Therefore, we consider systems, where the rate function can be expressed as in \cref{eq:crn_rate_fun} with mass action kinetics as in \cref{eq:mass_action_hazard}.
Throughout we will assume an exponential family form for the variational distribution as 
\begin{equation*}
    q(x\mid \theta)=h(x)\exp(\theta^\top s(x)-A(\theta)),
\end{equation*}
with base measure $h:\, \mathcal X \rightarrow \mathbb R_{\geq 0}$, sufficient statistics $s:\, \mathcal X \rightarrow \mathbb R^p$ and log-partion function $A:\, \Theta \rightarrow \mathbb R$.
Note, that $F(\theta)=\nabla_\theta \nabla_\theta^\top A(\theta)$ and $\nabla_\theta \log q(x\mid \theta)=s(x)-\E_{q(x\mid \theta)}[s(X)]$. 
As an instantiation we use a product Poisson distribution
\begin{align*}
    q(x\mid \theta)=\prod_{i=1}^n \PoisDis(x_i \mid \exp \theta_i).
\end{align*}
Hence, we have base-measure $h(x)=\prod_{i=1}^n \frac{1}{x_i!}$, sufficient statistics $s(x)=x$, natural parameters or log rate parameters $\theta=[\theta_1,\dots,\theta_n]^\top$ and log-partition function $A(\theta)=\sum_{i=1}^n \exp(\theta_i)$.
For the measurements we assume a linear Gaussian measurement model for the latent state, \ie, we assume the observation likelihood
    $p(y_i \mid x) = \NDis(y_i \mid H x, \Sigma)$,
with the $i$th observation $y_i \in \mathbb R^m$, the observation model matrix $H \in \mathbb R^{m \times n}$, and observation noise covariance $\Sigma \in \mathbb R^{m \times m}$ .
Note, that the following derivations can also be applied to general non-linear observation models $H(x)$, by following a linearization procedure, as in extended Kalman filtering approaches, see, \eg, \citep{sarkka2013bayesian}.
\subsection{Approximate Filtering and Smoothing}
For approximate inference, we first consider the initialization step, see \cref{eq:initial_condition_approx_filter}. 
Here, we assume that $p_0(x)=\prod_{i=1}^n \PoisDis(x_i \mid \exp(\theta_{i,0}))$, with given initial log rate parameters $\{\theta_{1,0},\dots, \theta_{n,0} \}$. 
Hence, \cref{eq:initial_condition_approx_filter} computes to the initial parameter 
$\theta(0)=[\theta_{1,0},\dots, \theta_{n,0}]^\top$.
Next, we compute the prediction step for the filtering distribution in \cref{eq:pred_step_approx_filter}.
For \acp{crn}, the product Poisson variational distribution leads to closed-form updates, see \cref{app:crn_filter}, for the prior drift as 
\begin{equation}
\label{eq:filter_crn}
    \frac{\diff}{\diff t} \theta(t) = F(\theta(t))^{-1}  \sum_{j=1}^k c_j \nu_j \exp\left(\sum_{i=1}^n \underline{\nu}_{ij} \theta_i(t)\right),
\end{equation}
with the inverse Fisher information matrix $F(\theta)^{-1}=\diag([\exp(-\theta_1),\dots, \exp(-\theta_n)]^\top)$.

At the observation time points, we compute the update from $\theta = \theta(t_i^-)$ to $\theta^\ast = \theta(t)$ according to
\cref{eq:update_step_approx_filter}.
Given the exponential family distribution for $q(x \mid \theta)$ this leads to a moment matching condition as
$
    \E_{q(x \mid \theta^\ast)}[s(X)]=\E_{p(x \mid y_i)}[s(X)]
$,
where $p(x \mid y_i)$
is the exact posterior distribution.
Since the Gaussian likelihood $p(y_i \mid x)$ is not conjugate to the product Poisson distribution $q(x \mid \theta)$, the computation of the exact posterior $p(x \mid y_i)$ is generally intractable.
We propose an additional moment matching scheme in order to get closed-form updates, see \cref{app:crn_update}.
The resulting mean of the posterior can then be written as
\begin{equation*}
\label{eq:mean_update_crn}
   m =  \exp \theta +  F(\theta) H^\top( H F(\theta) H^\top + \Sigma )^{-1}(y_i-H \exp \theta),
\end{equation*}
where the exponential function operating on vectors is applied component wise. Note that the posterior mean can be negative, as such we truncate it at a small value $0<\epsilon_{\lambda} \ll 1$. Hence, the full approximate update in the natural parameter space can be written as
\begin{equation}
\begin{aligned}
\label{eq:site_update_crn}
    \xi_i=\log \left(\max \left\{m, \epsilon_{\lambda} \right\} \right) - \theta,
\end{aligned}
\end{equation}
where we set $\epsilon_{\lambda}=10^{-6}$.

Given the filtering distribution we can then find a closed form expression for the backward smoothing step in \cref{eq:smoothing_step_approx_filter} as 
\begin{equation}
\begin{aligned}
\label{eq:smoother_crn}
    \frac{\diff}{\diff t} \tilde \theta(t)=  F(\tilde \theta(t))^{-1} \sum_{j=1}^k & c_j \nu_j \exp \left(\sum_{i=1}^n \underline{\nu}_{ij} \tilde \theta_i(t)\right)\\
    &\cdot \exp\left(\sum_{i=1}^n \nu_{ij} (\tilde \theta_i(t)-\theta_i(t))\right),
    \end{aligned}
\end{equation}
for the derivation see \cref{app:crn_smoother}.
\subsection{Expectation Propagation for Latent Chemical Reaction Networks}
Finally, for \ac{ep} we update the site parameters as described in \cref{eq:site_mjp_update}. Specifically, $\tilde \xi_i^{(j)}$ is obtained from \cref{eq:site_update_crn} by substituting $\theta=\tilde \theta_{\neg i}^{(j)}$, the cavity parameters computed using \cref{eq:cavity_mjp_update}. This step corresponds to computing the tilted distribution, projecting it back to the parametric family, and subtracting the cavity parameters. We summarize our method in \cref{alg:ep_for_crns}.

\begin{algorithm}
\SetKwInOut{Output}{output}
\KwIn{Time Horizon $T$, observations $\{y_1,\dots, y_N\}$, observation times $\{t_1,\dots, t_N\}$, initial parameters $\theta(0)$, \ac{ep} iterations $K$, learning rate parameter $\epsilon$}
\KwOut{Smoother parameter $\tilde\theta
$, site parameter $\xi_i$}
Initialize sites $\xi_i=0 $\\
\For{$j \gets 1$ \KwTo $K$ }{
\tcp{Compute the filter}
Initialize time $t=0$\\
\For{$i \gets 1$ \KwTo $N$ }{
Solve filter ODE \labelcref{eq:filter_crn} from $t$ to $t_i$ with initial condition $\theta(t)$\\
Update the filter at $\theta(t_i)$ via \cref{eq:update_step_approx_filter_sites}\\
Update time $t \gets t_i$\\
}
Solve filter ODE \labelcref{eq:filter_crn} from $t_N$ to $T$ with initial condition $\theta(t)$\\
\tcp{Compute the smoother}
Solve smoother ODE \labelcref{eq:smoother_crn} from $T$ to $0$ with initial condition $\tilde\theta(T)=\theta(T)$\\
\tcp{Expectation Propagation Steps}
Compute the cavity parameter $\tilde \theta_{\neg i}$ via \cref{eq:cavity_mjp_update}\\
Compute the new site parameters $\tilde\xi_i$ via \cref{eq:site_update_crn} by setting  $\theta=\tilde \theta_{\neg i}$\\
Perform a damped update of the site parameters $\xi_i$ via \cref{eq:damped_site_mjp_update}
}
\caption{Expectation propagation for CRNs}
\label{alg:ep_for_crns}
\end{algorithm}

\subsection{Parameter Learning}
For the given approximation of a \ac{crn} model, we can find closed form solutions for the \ac{mstep} of the \ac{em} algorithm.
Here, we consider the initial parameters $\{\theta_{1,0},\dots, \theta_{n,0}\}$, the effective rate parameters $\{c_{1},\dots, c_k\}$ and the observation model parameters $H$ and $\Sigma$, \ie, $\phi=\{\{\theta_{i,0}\}_{i=1}^n, \{c_{j}\}_{j=1}^k, H, \Sigma\}$.
In \cref{app:param_learning_crn} we show how to compute the bound in \cref{eq:marginal_ll_bound} \wrt the individual parameters $\phi$ in closed form.
Additionally, we derive closed-form coordinate-wise update schemes for the individual parameters $\phi$.


\section{EXPERIMENTS}
We evaluate our proposed method across various instantiations of \acp{crn} and compare the results against multiple baseline approaches. In this section, we highlight the latent state inference tasks for both a Lotka-Volterra model and a motility model. Detailed information on the experiments, additional benchmarks, and insights into the parameter learning task can be found in \cref{sec:app_experiments}.
\paragraph{Lotka-Volterra Model.}
The Lotka-Volterra or predator-prey model, see, \eg, \citep{wilkinson2018stochastic}, is one of the most well-known models in population dynamics.
It describes the dynamic evolution of a prey species $\mathsf X_1$  and a predator species $\mathsf X_2$.
The model can be described by three reactions as
\begin{equation*}
\begin{aligned}
 &\mathsf{X}_1 \xrightarrow{c_1} 2 \mathsf{X}_1,
 &&\mathsf{X}_1 + \mathsf X_2 \xrightarrow{c_2} 2 \mathsf{X}_2,
  &&&\mathsf{X}_2 \xrightarrow{c_3} \emptyset.
\end{aligned}
\end{equation*}
\begin{figure}
    \centering
    \includegraphics{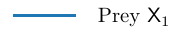} 
    \includegraphics{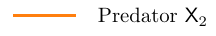}\\ 
    \includegraphics{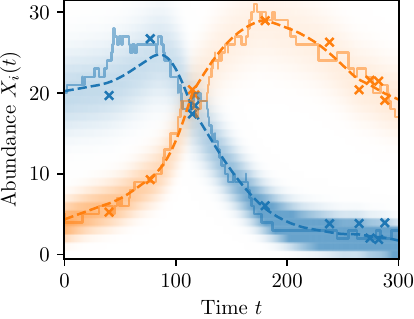}
    \caption{
    Simulation and latent state inference of a Lotka-Volterra model. Solid lines represent the ground truth trajectory, while crosses indicate the observations. Approximate inference results of our method are visualized with dashed lines for the posterior mean using and the background indicates the inferred marginal state probabilities.}
    \label{fig:lv}
\end{figure}
These reactions reflect the prey growth, the predator reproduction and the predator decline, with rates $c_1$, $c_2$, and $c_3$, respectively.
We model the reaction network system using \iac{mjp} with mass-action kinetics as in \cref{eq:crn_rate_fun,eq:mass_action_hazard}. 
We simulate the system using the Doob-Gillespie algorithm \citep{doob1945markoff,gillespie1976general} and consider that both species are observed at non-equidistant discrete time points subject to independent Gaussian noise. 
For inference, we exploit the discussed entropic matching method with \ac{ep}, using a product Poisson approximation.

We evaluate the performance on inferring the latent state path $X_{[0,T]}$ conditioned on the observations $Y_1,\dots, Y_N$. 
\cref{fig:lv} depicts the qualitative results of our method for a sample path with $10$ observations.
We notice that the inferred posterior mean closely tracks the qualitative behavior of the system dynamics. 
Additionally, the Poisson approximation gives a measure of uncertainty, indicated in the plot as the background.

For quantitative analysis, we infer the approximate posterior for $100$ sample trajectories and compare our results to various baselines, namely: \begin{inlineitemize}
\item a single \ac{ffbs} iteration using the entropic matching method without \ac{ep},
\item a Gaussian \ac{ads} based on the chemical Langevin equation, similar to the method described by \citet{cseke2016expectation},
\item the moment-based \ac{vi} method proposed by \citet{wildner2019moment}, and
\item an exact smoothing algorithm based on a truncation of the system, which serves as ground truth for our comparison.
\end{inlineitemize}
\begin{table}[h!]
\centering
\caption{Mean squared error in posterior mean averaged over trajectories and time}
\begin{tabular}{|c|c|c|c|}
\hline
\ac{ep} (Ours) &\ac{ffbs} Entropic &G \ac{ads} &MB\ac{vi}\\ 
\hline
\textbf{0.4581} &2.1989 & 1.9671& 1.6951\\ 
\hline
\end{tabular}
\label{table_lv_results}
\end{table}
\cref{table_lv_results} shows the mean squared error in the posterior mean of the approximate methods compared to the exact posterior mean of the truncated system.
Our method demonstrates superior performance overall. The improvement achieved with the \ac{ep} algorithm is substantial compared to the single \ac{ffbs} iteration with entropic matching.  Our method also outperforms the Gaussian \ac{ads}, which suffers from approximation errors due to modeling the \ac{mjp} as an \ac{sde}. This underscores the benefit of directly modeling the system as an \ac{mjp}, especially for low populations. Finally, the moment-based \ac{vi} method optimizes the rates describing the posterior process, whereas our method optimizes the parameters of the posterior marginal distributions directly, which leads to better mean estimates.

\paragraph{Motility Model.}

\begin{figure*}
\centering \includegraphics{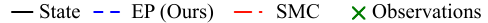}\\
\includegraphics{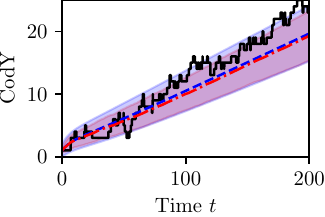}\hfill
\includegraphics{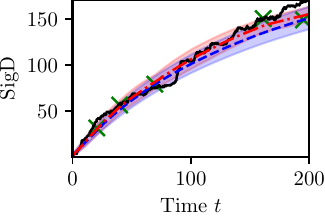}\hfill
\includegraphics{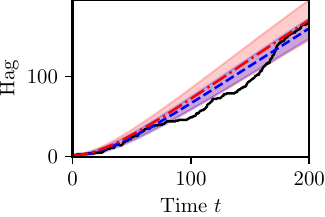}\hfill
    \caption{Three selected species from the motility model adapted from \citep{wilkinson2010parameter}, where only the species $\text{SigD}$ is noisily observed (crosses). The plots compare the results of our method and \iac{smc} approach by indicating their resulting mean and standard deviation.}
    \label{fig:wilk}
\end{figure*}
As a second more challenging example, we consider the motility model for bacterial gene regulation introduced by \citet{wilkinson2010parameter}. 
The model consists of a total of nine species and twelve reactions as 
\begin{equation*}
\begin{gathered}
\begin{aligned}
 &\text{codY} \xrightarrow{c_1} \text{codY} + \text{CodY},
 &&\text{CodY} \xrightarrow{c_2} \emptyset,
 \end{aligned}\\
  \begin{aligned}
 &\text{flache} \xrightarrow{c_3} \text{flache} + \text{SigD},
&&\text{SigD} \xrightarrow{c_4}\emptyset,
\end{aligned}\\
\begin{aligned}
 &\text{SigD\_hag} \xrightarrow{c_5} \text{SigD} + \text{hag} +\text{Hag},
 &\text{Hag} \xrightarrow{c_6} \emptyset,
 \end{aligned}\\
  \text{SigD} +\text{hag} \xrightarrow{c_7} \text{SigD\_hag}, \hspace{0.48em}
   \text{SigD\_hag}  \xrightarrow{c_8} \text{SigD} +\text{hag},\\
 \text{CodY} +\text{flache} \xrightarrow{c_9} \text{CodY\_flache},\\
  \text{CodY\_flache}\xrightarrow{c_{10}} \text{CodY} + \text{flache},\\
\!\quad\text{CodY} +\text{hag} \xrightarrow{c_{11}} \text{CodY\_hag},\\
\!\quad\text{CodY\_hag} \xrightarrow{c_{12}} \text{CodY} + \text{hag}.
\end{gathered}
\end{equation*}

In this setup, we only have access to noisy observations of $\text{SigD}$. 
The exact smoothing algorithm based on truncating the state space, used as ground truth in the Lotka-Volterra model, becomes intractable for higher-dimensional models like the motility model due to the exponential growth of the state space. This limitation underscores the necessity for approximate methods.
We again exploit a product Poisson approximation for the posterior distribution and estimate the latent state. 
To validate our approach we compare the results with \iac{smc} approach, with a sufficiently large number of samples $N_s=10000$, which acts as a reliable approximation of the ground truth.
We did not compare to the Gaussian \ac{ads} due to numerical instability, nor to the moment-based \ac{vi} method, as the number of moment equations that must be manually derived grows quadratically with the number of species. \cref{fig:wilk} shows qualitative results for three of the nine species given a sample trajectory.
The additional trajectories can be found in \cref{sec:app_experiments}. 

We observe that the trajectories for the noisily observed $\text{SigD}$, as well as the fully latent $\text{Hag}$ and $\text{CodY}$, can be tracked accurately. Our approximate solution yields results that are closely aligned with those obtained using the \ac{smc} approach, which serves as the ground truth. This demonstrates that our method provides reliable latent state inference while offering a more practical and scalable alternative for complex models.
\section{CONCLUSION}
We presented a principled inference framework for \acp{mjp} based on an entropic matching method embedded into an \ac{ep} algorithm. 
The new method arrives at closed-form results for the important class of \acp{crn} used within the field of systems biology.
The analytic nature of the closed-form message passing scheme makes the method highly scalable and fast. 

One limitation of our work is the expressiveness of the product Poisson approximation. As it only supplies one degree of freedom per species, it can only model an increase in variance by an increase in mean. Nevertheless, even with this limited variational distribution, our method demonstrates strong performance. This highlights the potential of our approach for addressing complex continuous-time Bayesian inference problems. Future work can build on this foundation by exploring more expressive variational distributions to further enhance the accuracy and applicability of the method.
A potential candidate could be an energy based model \citep{du2019implicit}, which would yield an expressive class for the approximate posterior distribution. This would go well beyond the presented setup of using a product form posterior distribution with only one parameter per dimension.
Though closed-form analytic updates might not be possible anymore, we think that advanced \ac{mcmc} methods \citep{sun2023discrete} could still lead to a scalable algorithm.
This would enable for rather complex observation likelihoods, which have for example also been discussed in the context of discrete-time systems \citep{johnson2016composing} and within the \ac{ep} framework \citep{vehtari2020expectation}. Moreover, we aim to extend our framework to other classes of \acp{mjp}, such as queueing systems, where entropic matching has recently been used for the filtering problem \citep{eich2024approx}.

\subsubsection*{Acknowledgements}
This work has been co-funded by the German Research Foundation (DFG) within the Collaborative Research Center (CRC) 1053 MAKI and project number 517777863 and by the Federal Ministry of Education and Research as part of the Software Campus project RL4MFRP (funding code 01IS23067).


\bibliographystyle{abbrvnat}
\bibliography{bibliography}

\section*{Checklist}



 \begin{enumerate}

 \item For all models and algorithms presented, check if you include:
 \begin{enumerate}
   \item A clear description of the mathematical setting, assumptions, algorithm, and/or model. [Yes]
   \item An analysis of the properties and complexity (time, space, sample size) of any algorithm. [Yes, see additional information on experiments in \cref{sec:app_experiments} ]
   \item (Optional) Anonymized source code, with specification of all dependencies, including external libraries. [Yes]
 \end{enumerate}

 \item For any theoretical claim, check if you include:
 \begin{enumerate}
   \item Statements of the full set of assumptions of all theoretical results. [Yes]
   \item Complete proofs of all theoretical results. [Yes]
   \item Clear explanations of any assumptions. [Yes]     
 \end{enumerate}

 \item For all figures and tables that present empirical results, check if you include:
 \begin{enumerate}
   \item The code, data, and instructions needed to reproduce the main experimental results (either in the supplemental material or as a URL). [Yes]
   \item All the training details (e.g., data splits, hyperparameters, how they were chosen). [Yes, see \cref{sec:app_experiments}]
         \item A clear definition of the specific measure or statistics and error bars (e.g., with respect to the random seed after running experiments multiple times). [Yes]
         \item A description of the computing infrastructure used. (e.g., type of GPUs, internal cluster, or cloud provider). [Yes, see \cref{sec:app_experiments}]
 \end{enumerate}

 \item If you are using existing assets (e.g., code, data, models) or curating/releasing new assets, check if you include:
 \begin{enumerate}
   \item Citations of the creator If your work uses existing assets. [Not Applicable]
   \item The license information of the assets, if applicable. [Not Applicable]
   \item New assets either in the supplemental material or as a URL, if applicable. [Not Applicable]
   \item Information about consent from data providers/curators. [Not Applicable]
   \item Discussion of sensible content if applicable, e.g., personally identifiable information or offensive content. [Not Applicable]
 \end{enumerate}

 \item If you used crowdsourcing or conducted research with human subjects, check if you include:
 \begin{enumerate}
   \item The full text of instructions given to participants and screenshots. [Not Applicable]
   \item Descriptions of potential participant risks, with links to Institutional Review Board (IRB) approvals if applicable. [Not Applicable]
   \item The estimated hourly wage paid to participants and the total amount spent on participant compensation. [Not Applicable]
 \end{enumerate}

 \end{enumerate}

\newpage
\appendix

\makeatletter
\renewcommand{\thesection}{\arabic{section}}
\makeatother

\onecolumn
\aistatstitle{Entropic Matching for Expectation Propagation of Markov Jump Processes: \\
Supplementary Materials}
\section{SOME NOTES ON EXACT FILTERING AND SMOOTHING}
\label{app:derivation_forward_smoothing}
To find the smoothing distribution $\tilde \pi_t(x)$ we have multiple options.
It can be shown \cite{huang2016reconstructing, anderson1983smoothing, pardoux1981non} that the smoothing distribution can be factorized as
\begin{equation}
\label{eq:app_smooth_factorization}
    \tilde \pi_t(x) =\frac{\pi_t(x) \beta_t(x) }{\sum_{x' \in \mathcal X} \pi_t(x') \beta_t(x')},
\end{equation}
where $\beta_t(x)$ is the backward-filtering distribution given by
$
    \beta_t(x) \coloneqq p(y_{(t,T]} \mid X(t)=x)
$,
where $y_{(t,T]}\coloneqq \{y_i \mid i \in \{1,\dots,N\}: t_i \in (t,T]\}$ are only the \enquote{future} observations.
Note that $ \tilde \pi_T(x)= \Prob(X(T)=x \mid y_{[0,T]})=\pi_T(x)$, hence we have $\beta_T(x)=1$.
The backward-filtering distribution in between observation time points follows the differential form of the backward Chapman-Kolmogorov equation \cite{gardiner1985handbook} and is updated at observation time points \cite{huang2016reconstructing}, hence, we have
\begin{equation*}
\begin{aligned}
        &\beta_T(x)=1, &&\frac{\diff}{\diff t} \beta_t(x)= - [\mathcal L_t^\dagger \beta_t](x), &&&\beta_{t_i^-}(x)=\frac{p(y_i \mid x) \beta_{t_i}(x)}{\sum_{x' \in \mathcal X} p(y_i \mid x') \beta_{t_i}(x')}.
\end{aligned}
\end{equation*}
 Note that $\beta_t(x)$ is càglàd in $t$.
 Therefore, we can compute the smoothing distribution $\tilde \pi_t(x)$, by first computing the filtering distribution  $\pi_t(x)$ forwardly in time, subsequently, computing the backward-filtering distribution  $\beta_t(x)$  backwardly in time and then use \cref{eq:app_smooth_factorization} to compute $\tilde \pi_t(x)$.
 
 Another way is to note that an initial condition for the smoothing distribution is given by $\tilde \pi_0(x)\propto p_0(x) \beta_0(x)$. 
 Differentiating \cref{eq:app_smooth_factorization} by time, then gives rise to the forward smoothing dynamics \cite{huang2016reconstructing,  anderson1983smoothing}
 \begin{equation}
 \label{eq:app_exact_forward_smoothing_dynamics}
 \begin{aligned}
      &\tilde \pi_0(x)=\frac{p_0(x) \beta_0(x)}{\sum_{x' \in \mathcal X} p_0(x') \beta_0(x')}, &&\frac{\diff}{\diff t} \tilde \pi_t(x) = [\bar{\mathcal L}_t\tilde \pi_t](x),
 \end{aligned}
 \end{equation}
 where the forward smoothing operator $\bar{\mathcal L}_t$ for an arbitrary test function $\phi$ and the forward smoothing rate function $\bar \Lambda(x,x',t)$ are given as
 \begin{equation}
 \label{eq:app_exact_forward_smoothing_rates}
 \begin{aligned}
     & [\bar{\mathcal L}_t \phi](x) \coloneqq \sum_{x' \neq x} \bar \Lambda(x',x,t) \phi(x') - \bar \Lambda(x,x',t) \phi(x), &&\bar \Lambda(x,x',t)= \Lambda(x,x',t) \frac{\beta_t(x')}{\beta_t(x)}.
 \end{aligned} 
 \end{equation}
 Hence, the smoothing distribution $\tilde \pi_t(x)$ can also be calculated by solving the backwards filtering distribution $\beta_t(x)$ backwardly in time and subsequently solving for the smoothing distribution $\tilde \pi_t(x)$ using the forward smoothing dynamics in \cref{eq:app_exact_forward_smoothing_dynamics}.

  \vfill
\section{DERIVATION FOR THE FILTERING AND SMOOTHING VARIATIONAL PARAMETERS}

\subsection{Entropic Matching for Filtering}
\label{app:derivation_filter}
Here, we find the variational parameters $\theta(t)$ by considering 
\begin{equation*}
    \theta(t+h)=\argmin_{\theta'} \KLof*{\pi_{t+h}(x)|q(x\mid \theta')},
\end{equation*}
where $\pi_{t+h}(x)$ is the filtering distribution that is propagated for a small time step $h$. 
We  get an recursive algorithm by considering that the filtering dsitrbution at time point $t$ has the initial distribution $q(x \mid \theta(t))$, \ie, we have
\begin{equation*}
    \begin{aligned}
        &\frac{\diff}{\diff t} \pi_t(x) = [\mathcal L_t \pi_t](x), &&\pi_t(x)=q(x \mid \theta(t)).
    \end{aligned}
\end{equation*}
Hence, we can write 
\begin{equation*}
    \pi_{t+h}(x) = q(x \mid \theta(t)) + h \mathcal L_t q(x \mid \theta(t)) + o(h),
\end{equation*}
with $\lim_{h \to 0} \nicefrac{o(h)}{h} =0$.
For the \ac{kl} divergence we compute
\begin{equation*}
\begin{multlined}
     \KLof*{\pi_{t+h}(x)|q(x\mid \theta')}\\
     \begin{aligned}
 &=\KLof*{q(x \mid \theta(t)) + h \mathcal L_t q(x \mid \theta(t)) + o(h)|q(x\mid \theta')}\\
     &=\E_{\pi_{t+h}(x)}\left[\log \frac{q(X \mid \theta(t)) + h \mathcal L_t q(X \mid \theta(t)) + o(h)}{q(X\mid \theta')}\right]\\
     &=\E_{\pi_{t+h}(x)}\left[ \log\left( q(X \mid \theta(t)) + h \mathcal L_t q(X \mid \theta(t)) + o(h)\right) - \log q(X\mid \theta') \right]
     \end{aligned}
\end{multlined}
\end{equation*}
Using a Taylor series around $h=0$, we can find that for some coefficients $a$ and $b$, we have 
\begin{equation*}
    \log(a+ bh +o(h))=\log a +h \frac{b}{a} +o(h).
\end{equation*}
Hence, we calculate
\begin{equation*}
\begin{multlined}
     \KLof*{\pi_{t+h}(x)|q(x\mid \theta')}   \\
     \begin{aligned}
      &=\E_{\pi_{t+h}(x)}\left[ \log\left( q(X \mid \theta(t)) + h \mathcal L_t q(X \mid \theta(t)) + o(h)\right) - \log q(X\mid \theta') \right]\\
     &=\E_{\pi_{t+h}(x)}\left[\log q(X \mid \theta(t) +h \frac{\mathcal L_t q(X \mid \theta(t))}{q(X \mid \theta(t)} +o(h) -  \log q(X\mid \theta')\right]\\
     &=\sum_{x \in \mathcal X} \pi_{t+h}(x) \left[\log q(x \mid \theta(t) +h \frac{\mathcal L_t q(x \mid \theta(t))}{q(x \mid \theta(t)} +o(h) -  \log q(x\mid \theta')\right]\\
     &
     \begin{multlined}
         =\sum_{x \in \mathcal X} \left(q(x \mid \theta(t)) + h \mathcal L_t q(x \mid \theta(t)) + o(h) \right)\\
     \cdot \left[\log q(x \mid \theta(t) +h \frac{\mathcal L_t q(x \mid \theta(t))}{q(x \mid \theta(t)} +o(h) -  \log q(x\mid \theta')\right]
     \end{multlined}\\
     &
     \begin{multlined}
          =\sum_{x \in \mathcal X} \left\{ q(x \mid \theta(t)) \log \frac{q(x \mid \theta(t))}{q(x\mid \theta')} \right.\\
      \left.+ h  \left[ q(x \mid \theta(t)) \frac{\mathcal L_t q(x \mid \theta(t))}{q(x \mid \theta(t))} +\mathcal L_t q(x \mid \theta(t)) \log \frac{q(x \mid \theta(t))}{q(x \mid \theta')}\right] + o(h) \right\}
     \end{multlined}\\
     &
     \begin{multlined}
         =\KLof*{q(x \mid \theta(t)) | q(x \mid \theta')}\\
     +h \E_{q(x \mid \theta(t))}\left[\frac{\mathcal L_t q(X \mid \theta(t))}{q(X \mid \theta(t)} + \frac{\mathcal L_t q(X \mid \theta(t))}{q(X \mid \theta(t))}\log \frac{q(X \mid \theta(t))}{q(X \mid \theta')}\right] +o(h).
     \end{multlined}\\
     \end{aligned}
\end{multlined}
\end{equation*}
By assuming that for small $h$ the parameter $\theta'=\theta(t+h)$ is close to the parameter $\theta=\theta(t)$, we can exploit a series expansion in $\theta -\theta'$ up to second order of the \ac{kl} divergence as
\begin{equation*}
    \KLof*{q(x \mid \theta) | q(x \mid \theta')}=\frac{1}{2} (\theta'-\theta)^\top F(\theta) (\theta'-\theta),
\end{equation*}
where $F(\theta)\coloneqq - \E_{q(x \mid \theta)}\left[\nabla_\theta \nabla_\theta^\top \log q(X \mid \theta)\right]$ is the Fisher information matrix.
Hence, we compute
\begin{equation*}
\begin{aligned}
        0&=\nabla_{\theta'} \KLof*{\pi_{t+h}(x)|q(x\mid \theta')}\vert_{\theta'=\theta(t+h)}\\
    &=F(\theta(t)) (\theta(t+h)-\theta(t))- h\E_{q(x \mid \theta(t))}\left[\nabla_{\theta'} \log q(X\mid \theta(t+h)) \frac{\mathcal L_t q(X \mid \theta(t))}{q(X \mid \theta (t))}\right] +o(h)
\end{aligned}
\end{equation*}
Dividing both sides by $h$ and taking the limit $h \to 0$ we obtain
\begin{equation*}
    \begin{aligned}
        0&= F(\theta(t)) \frac{\diff}{\diff t} \theta(t) -\E_{q(x \mid \theta(t))}\left[ \nabla_\theta \log q(X \mid \theta(t)) \frac{\mathcal L_t q(X \mid \theta(t))}{q(X \mid \theta (t))} \right]\\
        &=F(\theta(t)) \frac{\diff}{\diff t} \theta(t) - \sum_x q(x \mid \theta(t)) \nabla_{\theta} \log q(x \mid \theta(t)) \frac{\mathcal L_t q(x \mid \theta(t))} {q(x \mid \theta (t))}\\
        &=F(\theta(t)) \frac{\diff}{\diff t} \theta(t) - \sum_x \nabla_{\theta} \log q(x \mid \theta(t)) \mathcal L_t q(x \mid \theta(t))\\
        &=F(\theta(t)) \frac{\diff}{\diff t} \theta(t) - \sum_x q(x \mid \theta(t)) \mathcal L_t^\dagger   \nabla_{\theta} \log q(x \mid \theta(t)) \\
        &=F(\theta(t)) \frac{\diff}{\diff t} \theta(t) - \E_{q(x \mid \theta(t))}\left[\mathcal L_t^\dagger   \nabla_{\theta} \log q(X \mid \theta(t))\right].
        \end{aligned}
\end{equation*}
This leads to the drift in between observation time points
 \begin{equation*}
     \frac{\diff}{\diff t} \theta(t)= F(\theta(t))^{-1} \E_{q(x \mid {\theta}(t))}\left[\mathcal L_t^\dagger \nabla_\theta \log q(X \mid \theta(t)) \right].
 \end{equation*}
 
At the observation time points we reset the parameter to 
 \begin{equation*}
     \theta(t_i)= \argmin_{\theta'} \KLof*{ Z^{-1}p(y_i\mid x) q(x\mid \theta(t_i^-))| q(x\mid \theta')},
 \end{equation*}
 where the first argument in the KL divergence corresponds to the posterior distribution obtained from Bayes' rule and $Z$ refers to the normalization constant. 
 We usually use the unnormalized posterior as first argument, which does not affect the optimization, as the normalization constant does not depend on $\theta'$.
 
The initial condition $\theta(0)=\theta_0$ is given by
  \begin{equation*}
     \theta_0= \argmin_{\theta'} \KLof*{ p_0(x)| q(x\mid \theta')}.
 \end{equation*}

 \subsection{Entropic Matching for Smoothing}
\label{app:derivation_smoothing}
 Similar to the filtering approximation, we use the entropic matching method backwards in time for the smoother
 \begin{equation*}
     \tilde \pi_t(x) \approx q(x \mid \tilde{\theta}(t)).
 \end{equation*}
 Here, we compute 
 \begin{equation*}
     \tilde \theta(t-h)=\argmin_{\tilde \theta} \KLof*{\tilde \pi_{t-h}(x) | q(x \mid \tilde \theta)}.
 \end{equation*}
 Further, we assume that
 \begin{equation*}
    \begin{aligned}
        &\frac{\diff}{\diff t} \tilde \pi_t(x) = -\tilde{\mathcal L}_t \tilde \pi_t(x), &&\tilde \pi_t(x)=q(x \mid \tilde \theta(t)).
    \end{aligned}
\end{equation*}
 For the smoothing distribution we can compute the evolution backwards in time by
 \begin{equation*}
     \tilde \pi_{t-h}(x)= q(x \mid \tilde \theta(t)) + h \tilde{\mathcal L}_t q(x \mid \tilde \theta(t)) + o(h).
 \end{equation*}
 Hence, similar to the derivation of the approximate filter we have
 \begin{equation*}
     \begin{aligned}
         0 &= F(\tilde \theta(t)) (\tilde \theta(t- h)- \tilde \theta(t))- h\E_{q(x \mid \tilde \theta(t))}\left[\nabla_{\tilde \theta} \log q(X\mid \tilde \theta(t - h)) \frac{\tilde{\mathcal L}_t q(X \mid \tilde \theta(t))}{q(X \mid \tilde\theta (t))}\right] +o(h).
     \end{aligned}
 \end{equation*}
Dividing both sides by $h$ and taking the limit $h \to 0$ we arrive analog to the above derivation at
 \begin{equation*}
     \frac{\diff}{\diff t} \tilde \theta(t)= - F(\tilde \theta(t))^{-1} \E_{q(x \mid \tilde{\theta}(t))}\left[\tilde{\mathcal L}_t^\dagger \nabla_\theta \log q(X \mid \tilde \theta(t)) \right],
 \end{equation*}
 with end-point condition $\tilde \theta(T)=\theta(T)$.

\section{PARAMETER LEARNING}
\label{app:param_learning_mjp}
For parameter learning we use an approximate \ac{em} scheme.
For the derivation of the \ac{em} algorithm, we consider that instead of maximizing the intractable marginal likelihood $p(y_{[0,T]} \mid \phi )$ \wrt some parameters $\phi$, a tractable lower bound $\mathrm L(\phi)$ as
\begin{equation*}
    \log p(y_{[0,T]} \mid \phi ) \geq \mathrm L(\phi) \coloneqq \E_{\Qrob}\left[ \log \frac{\diff \Prob}{\diff \Qrob} (X_{[0,T]}, y_{[0,T]} , \phi) \right],
\end{equation*}
where $\frac{\diff \Prob}{\diff \Qrob}$ is the Radon-Nikodym derivative of the path measure $\Prob(X_{[0,T]}\in \cdot, Y_{[0,T]} \in \cdot \mid \phi)$ for the latent paths $X_{[0,T]} \coloneqq \{X(t) \mid t \in [0,T]\}$ and all observations $Y_{[0,T]}$ \wrt an an approximating probability measure $\Qrob(X_{[0,T]} \in \cdot)$ over latent paths and the Lebesgue measure over all observations $Y_{[0,T]}$.
Optimizing this bound \wrt the approximate probability measure yields the exact posterior probability measure $\Qrob(X_{[0,T]} \in \cdot)= \Prob(X_{[0,T]} \in \cdot \mid y_{[0,T]}, \phi)$.
We note that the bound $\mathrm L(\phi)$ can be written as
\begin{equation*}
    \begin{aligned}
        \mathrm L(\phi)&=\E_{\Qrob}\left[ \log \frac{\diff \Prob}{\diff \Qrob} (X_{[0,T]}, y_{[0,T]} , \phi) \right]\\
        &=\E_{\Qrob}\left[ \log \left(p( y_{[0,T]} \mid X_{[0,T]}, \phi) \frac{\diff \Prob}{\diff \Qrob}(X_{[0,T]}, \phi)  \right) \right]\\
        &=\E_{\Qrob}\left[ \log \frac{\diff \Prob}{\diff \Qrob}(X_{[0,T]}, \phi) \right] + \sum_{i=1}^N \E_{q(x, t_i)}\left[\log p(y_i \mid X, \phi)\right]\\
        &=- \KLof*{\Qrob(X_{[0,T]} \in \cdot) | \Prob(X_{[0,T]} \in \cdot \mid \phi)} + \sum_{i=1}^N \E_{q(x, t_i)}\left[\log p(y_i \mid X, \phi)\right],
    \end{aligned}
\end{equation*}
where we denote by $q_t(x)\coloneqq \Qrob(X(t)=x)$ the time point-wise marginal of the approximate path measure.
Note that both the approximate path measure $\Qrob(X_{[0,T]} \in \cdot)$ and the prior path measure  $\Prob(X_{[0,T]} \in \cdot \mid \phi)$ are path measures induced by \iac{mjp}. 
For the prior path measure $\Prob(X_{[0,T]} \in \cdot \mid \phi)$ this is easy to note, since $\{X(t)\}$ is \iac{mjp}.
For the approximate posterior path-measure $\Qrob(X_{[0,T]} \in \cdot )=\Prob(X_{[0,T]} \in \cdot \mid y_{[0,T]}, \phi)$ this can be seen, as the equations for its time point-wise marginals $q_t(x)=\Qrob(X(t)=x)=\Prob(X(t)=x \mid y_{[0,T]}, \phi)=\tilde \pi_t(x)$ are given by the evolution of the smoothing distribution in \cref{eq:app_exact_forward_smoothing_dynamics}.
This is evolution equation has the form of a master equation with the rate function $\bar \Lambda(x,x',t)$ as defined in \cref{eq:app_exact_forward_smoothing_rates}, and therefore, the posterior process is also \iac{mjp}.
The \ac{kl} divergence for two \acp{mjp}, which is the expected log Radon-Nikodym derivative, has been derived in the literature multiple times, see, \eg, \cite{kipnis1998scaling, opper2007variational, hanson2007applied, cohn2010mean}.
The \ac{kl} divergence computes to
\begin{equation*}
\begin{multlined}
       \KLof*{\Qrob(X_{[0,T]} \in \cdot) | \Prob(X_{[0,T]} \in \cdot \mid \phi)}=\KLof*{q(x,0)| p_0(x \mid \phi)} \\
        + \int_{0}^T \E_{q_t(x)}\left[ \sum_{x' \neq X} \bar \Lambda(X,x',t) \log \frac{\bar \Lambda(X,x',t)}{\Lambda(X,x',t \mid \phi)} - \left(\bar \Lambda(X,x',t) - \Lambda(X,x',t\mid \phi)\right) \right] \diff t,
\end{multlined}
\end{equation*}
where only the prior rates $\Lambda(x,x',t \mid \phi)$ depend on the parameters $\phi$. 
Note, that the forward posterior rates $\bar \Lambda(x,x',t)$ in \cref{eq:app_exact_forward_smoothing_rates} can be written in terms of the filtering and smoothing distribution by exploiting \cref{eq:app_smooth_factorization} as
\begin{equation*}
    \bar \Lambda(x,x',t)= \Lambda(x,x',t) \frac{\beta_t(x')}{\beta_t(x)}= \Lambda(x,x',t) \frac{\tilde \pi_t(x')}{\tilde \pi_t(x)} \frac{\pi_t(x)}{\pi_t(x')}.
\end{equation*}
Therefore, we arrive at the following expression for the bound as 
\begin{equation*}
    \begin{multlined}
        \mathrm L(\phi)= \E_{q_0(x)}\left[\log \frac{ p_0(X \mid \phi)}{q_0(X)} \right] 
        + \int_{0}^T \E_{q_t(x)}\left[ \sum_{x' \neq X} \Lambda(X,x',t) \frac{\tilde \pi_t(x')}{\tilde \pi_t(X)} \frac{\pi_t(X)}{\pi_t(x')} \right.\\
        \cdot \log \frac{ \Lambda(X,x',t \mid \phi)}{\Lambda(X,x',t)} - \Lambda(X,x',t\mid \phi) \Bigg] \diff t
        + \sum_{i=1}^N \E_{q(x, t_i)}\left[\log p(y_i \mid X, \phi)\right]+\text{const}.
    \end{multlined}
\end{equation*}

Since, the computation $\Qrob(X_{[0,T]} \in \cdot)= \Prob(X_{[0,T]} \in \cdot \mid y_{[0,T]}, \phi)$ of the exact posterior probability measure is intractable, we replace its time-point wise marginals with the approximate filtering and smoothing distribution as $p(x,t \mid y_{[0,T]}, \phi) = \tilde \pi_t (x) \approx q(x \mid \tilde \theta(t))$ and $\pi_t (x)\approx q(x \mid \theta(t))$, respectively. 
We keep these distributions fixed and this yields an expression for the bound as
\begin{equation}
\label{eq:app_elbo}
    \begin{split}
        \mathrm L(\phi)=\E_{q(x \mid \tilde \theta(0)))}\left[\log \frac{ p_0(X \mid \phi)}{q(X \mid \tilde \theta(0))}\right] 
        + \int_{0}^T \E_{q(x \mid \tilde \theta(t))}\left[ \sum_{x' \neq X} \Lambda(X,x',t) \frac{q(x' \mid \tilde \theta(t))}{q(X \mid \tilde \theta(t))} \frac{q(X\mid  \theta(t))}{q(x' \mid \theta(t))} \right. \\
         \cdot \log \frac{\Lambda(X,x',t \mid \phi)}{\Lambda(X,x',t)}  - \Lambda(X,x',t\mid \phi) \Bigg] \diff t
        + \sum_{i=1}^N \E_{q(x \mid \tilde \theta(t_i))}\left[\log p(y_i \mid X, \phi)\right]+\text{const}.
    \end{split}
\end{equation}
Therefore, we can find the optimal parameters, by iteratively computing the filtering and smoothing distribution $\{q(x \mid \theta(t)) \mid t \in [0,T]\}$ and $\{q(x \mid \tilde \theta(t))\mid t \in [0,T]\}$, respectively, and optimizing \cref{eq:app_elbo} \wrt the parameters $\phi$.

\section{DERIVATION FOR CHEMICAL REACTION NETWORKS}
\subsection{Filtering Distribution}
\label{app:crn_filter}
The evolution of the approximate filtering distribution is given by
 \begin{equation*}
     \frac{\diff}{\diff t} \theta(t)= F(\theta(t))^{-1} \E_{q(x \mid {\theta}(t))}\left[\mathcal L_t^\dagger \nabla_\theta \log q(X \mid \theta(t)) \right].
 \end{equation*}
 We assume a product Poisson variational distribution $q(x \mid {\theta})$ and the adjoint operator $\mathcal L_t^\dagger$ is characterized by \ac{crn} dynamics with mass action kinetics. 
 Therefore, we compute
 \begin{equation*}
     \begin{aligned}
          &\E_{q(x \mid {\theta}(t))}\left[\mathcal L_t^\dagger \nabla_\theta \log q(X \mid \theta(t)) \right]\\
          &{}= \E_{q(x \mid {\theta}(t))}\left[ \sum_{x' \neq X} \Lambda(X, x' ,t) \left( \nabla_\theta \log q(x' \mid \theta(t)) - \nabla_\theta \log q(X \mid \theta(t)) \right) \right]\\
          &{}= \E_{q(x \mid {\theta}(t))}\left[ \sum_{x' \neq X} \Lambda(X, x' ,t) \left(x' -X \right) \right]\\
          &{}=\E_{q(x \mid {\theta}(t))}\left[ \sum_{x' \neq X}  \sum_{j=1}^k \1(x'=X + \nu_j)\lambda_j(X)  \left(x' -X \right) \right]          \\
          &{}=\E_{q(x \mid {\theta}(t))}\left[  \sum_{j=1}^k  \nu_j\lambda_j(X)   \right]   =  \sum_{j=1}^k  \nu_j  \E_{q(x \mid {\theta}(t))} \left[\lambda_j(X)   \right] \\
           &{}=\sum_{j=1}^k  c_j \nu_j \E_{q(x \mid {\theta}(t))}\left[ \prod_{i=1}^{n} (X_i)_{\underline{\nu}_{ij}}  \right]  =      \sum_{j=1}^k   c_j \nu_j \prod_{i=1}^{n}  \left(\exp \theta_i(t)\right)^{\underline{\nu}_{ij}}
     \end{aligned}
 \end{equation*}
 The last line is computed, by noting that the Poisson random variables are independent and the $m$th factorial moment of a Poisson random variable with mean $\lambda$ is given as $\E[(X)_m]=\lambda^m$. 
 By writing this results in the natural parameterization, we arrive at 
 \begin{equation*}
     \frac{\diff}{\diff t} \theta(t)= F(\theta(t))^{-1}  \sum_{j=1}^k   c_j  \nu_j\exp(\sum_{i=1}^n \underline{\nu}_{ij} \theta_i(t)).
 \end{equation*}
 
 \subsection{Kalman-type Updates for the Filtering Distribution}
 \label{app:crn_update}

At the observation time points, we have discrete-time resets from $\theta=\theta(t_i^-)$ to $\theta^\ast = \theta(t)$as
 \begin{equation*}
 \begin{aligned}
      &\theta^\ast= \argmin_{\theta'} \KLof*{ p(y_i\mid x) q(x\mid \theta)| q(x\mid \theta')}, &&\forall i = 1,\dots, N
 \end{aligned}
 \end{equation*}
The optimal new parameter $\theta^\ast = \theta(t)$ can be found by computing the derivative of the \ac{kl} divergence and setting it to zero.
Given the exponential family distribution for $q(x \mid \theta)$ this leads to a moment matching condition as
$
    \E_{q(x \mid \theta^\ast)}[s(X)]=\E_{p(x \mid y_i)}[s(X)]
$,
where $p(x \mid y_i)
  =\nicefrac{p(y_i \mid x) q(x \mid \theta)}{\sum_{x'} p(y_i \mid x') q(x' \mid \theta)}$ 
is the exact posterior distribution.
Since the Gaussian likelihood $p(y_i \mid x)$ is not conjugate to the product Poisson distribution $q(x \mid \theta)$, the computation of the exact posterior $p(x \mid y_i)$ is generally intractable.
However, we can find a conjugate update, by assuming a Gaussian distribution at the observation time points as $q(x \mid \theta) \approx \NDis(x\mid m, P)$.
By first performing a minimization of the \ac{kl} divergence $\KLof*{q(x \mid \theta) | \NDis(x \mid m, P)}$, \wrt the parameters $m$ and $P$ of the Gaussian distribution, we can find an approximation to the product Poisson distribution $q(x \mid \theta)$.
This yields the standard Gaussian approximation of the Poisson distribution, with parameters $m=[\exp \theta_1,\dots, \exp \theta_n]^\top$ and $P=\diag([\exp \theta_1,\dots, \exp \theta_n]^\top)$
This then leads to conjugate updates, as for the new mean parameter $m^\ast=\E_{p(x \mid y_i)}[X]$ in the update equations for the Kalman filter, see, \eg, \cite{sarkka2013bayesian}, as
\begin{equation*}
      m^\ast=m + P H^\top(H P H^\top + \Sigma )^{-1}(y_i-H m).
\end{equation*}
Finally we can compute the optimal variational parameters after the jump according to the moment matching condition $\E_{q(x \mid \theta^\ast)}[s(X)]=\E_{p(x \mid y_i)}[s(X)]$.
Note, that the mean of the Gaussian can be negative, as such we truncate it at a small value $0<\epsilon_{\lambda} \ll 1$. This yields
\begin{equation*}
    \theta^\ast=\log \left(\max \left\{\exp \theta +  F(\theta) H^\top( H F(\theta) H^\top + \Sigma )^{-1}(y_i-H \exp \theta), \epsilon_{\lambda} \right\} \right).
\end{equation*}
The resulting update therefore can be written as
\begin{equation*}
    \xi_i=\log \left(\max \left\{\exp \theta +  F(\theta) H^\top( H F(\theta) H^\top + \Sigma )^{-1}(y_i-H \exp \theta), \epsilon_{\lambda} \right\} \right)-\theta.
\end{equation*} 
 \subsection{Smoothing Distribution}
\label{app:crn_smoother}
For the derivation to the smoothing distribution we compute
 \begin{equation*}
     \frac{\diff}{\diff t} \tilde \theta(t)= - F(\tilde \theta(t))^{-1} \E_{q(x \mid \tilde{\theta}(t))}\left[\tilde{\mathcal L}_t^\dagger \nabla_\theta \log q(X \mid \tilde \theta(t)) \right].
 \end{equation*}
 The expectation on the \rhs computes to 
 \allowdisplaybreaks{
 \begin{align*}
      &\E_{q(x \mid \tilde{\theta}(t))}\left[\tilde{\mathcal L}_t^\dagger \nabla_\theta \log q(X \mid \tilde \theta(t)) \right] =\E_{q(x \mid \tilde{\theta}(t))}\left[ \sum_{x' \neq X} \tilde \Lambda(X, x' ,t) \left(x' -X \right) \right]\\
      &= \E_{q(x \mid \tilde{\theta}(t))}\left[ \sum_{x' \neq X}  \Lambda( x', X ,t) \frac{\pi_t(x')}{\pi(X, t)} \left(x' -X \right) \right]\\
      &= \E_{q(x \mid \tilde{\theta}(t))}\left[ \sum_{x' \neq X}  \Lambda( x', X ,t) \frac{q(x' \mid \theta(t))}{q(X \mid \theta(t))} \left(x' -X \right) \right]\\
      &= \E_{q(x \mid \tilde{\theta}(t))}\left[ \sum_{x' \neq X}  \sum_{j=1}^k \1(X=x' + \nu_j)\lambda_j(x')  \left\{\prod_{i=1}^n \frac{X_i!}{x_i'!} \exp((x_i'-X_i) \theta_{i}(t))\right\} \left(x' -X \right) \right]\\
      &=\E_{q(x \mid \tilde{\theta}(t))}\left[ \sum_{j=1}^k \lambda_j(X-\nu_j)  \left\{\prod_{i=1}^n \frac{X_i!}{(X_i-\nu_{ij})!} \exp(-\nu_{ij} \theta_{i}(t))\right\} \left(-\nu_j\right) \right]\\
      &=\E_{q(x \mid \tilde{\theta}(t))}\left[ \sum_{j=1}^k \lambda_j(X-\nu_j)  \left\{\prod_{i=1}^n (X_i)_{\nu_{ij}} \right\} \exp\left(-\sum_{i=1}^n \nu_{ij} \theta_{i}(t)\right) \left(-\nu_j\right) \right]\\
      &=- \sum_{j=1}^k  c_j  \nu_j\exp\left(-\sum_{i=1}^n \nu_{ij} \theta_{i}(t)\right)  \E_{q(x \mid \tilde{\theta}(t))}\left[ \left\{\prod_{i=1}^n (X_i-\nu_{ij})_{\nu_{ij}} \right\}  \left\{\prod_{i=1}^n (X_i)_{\nu_{ij}} \right\} \right]\\
            &=- \sum_{j=1}^k   c_j \nu_j \exp\left(-\sum_{i=1}^n \nu_{ij} \theta_{i}(t)\right)  \E_{q(x \mid \tilde{\theta}(t))}\left[ \prod_{i=1}^n (X_i-\nu_{ij})_{\underline{\nu}_{ij}} (X_i)_{\nu_{ij}} \right]\\
            &=- \sum_{j=1}^k   c_j \nu_j \exp\left(-\sum_{i=1}^n \nu_{ij} \theta_{i}(t)\right)  \E_{q(x \mid \tilde{\theta}(t))}\left[ \prod_{i=1}^n (X_i-\nu_{ij})_{\underline{\nu}_{ij}} (X_i)_{\nu_{ij}} \right]\\
            &=- \sum_{j=1}^k   c_j \nu_j\exp\left(-\sum_{i=1}^n \nu_{ij} \theta_{i}(t)\right)  \E_{q(x \mid \tilde{\theta}(t))}\left[ \prod_{i=1}^n (X_i)_{\nu_{ij} +\underline{\nu}_{ij}}  \right]\\
            &=- \sum_{j=1}^k   c_j \nu_j \exp\left(-\sum_{i=1}^n \nu_{ij} \theta_{i}(t)\right) \prod_{i=1}^n \lambda_i(t)^{\nu_{ij} +\underline{\nu}_{ij}}\\
            &=- \sum_{j=1}^k  c_j \nu_j \exp\left(-\sum_{i=1}^n \nu_{ij} \theta_{i}(t)\right) \exp\left(\sum_{i=1}^n (\nu_{ij} +\underline{\nu}_{ij}) \tilde \theta_i(t)\right)\\
            &=- \sum_{j=1}^k  c_j  \nu_j  \exp \left(\sum_{i=1}^n \underline{\nu}_{ij} \tilde \theta_i(t)\right) \exp\left(\sum_{i=1}^n \nu_{ij} (\tilde \theta_i(t)-\theta_i(t))\right).
 \end{align*}}
 This yields the smoothing dynamics as 
 \begin{equation*}
    \frac{\diff}{\diff t} \tilde \theta(t)=  F(\tilde \theta(t))^{-1} \sum_{j=1}^k c_j \nu_j \exp \left(\sum_{i=1}^n \underline{\nu}_{ij} \tilde \theta_i(t)\right) \exp\left(\sum_{i=1}^n \nu_{ij} (\tilde \theta_i(t)-\theta_i(t))\right).
\end{equation*}

\section{PARAMETER LEARNING FOR LATENT CHEMICAL REACTION NETWORKS}
\label{app:param_learning_crn}
Here, we give an expression for the bound $\mathrm L(\phi)$ for the case of a product Poisson approximation for the filtering and smoothing distributions, for dynamics given as \iac{crn} with mass action kinetics, and a linear Gaussian observation model. 
For the parameters we consider the initial condition parameters $\{\theta_{i,0}\}_{i=0}^n$, the effective rate parameters $\{c_j\}_{j=1}^k$, and the linear observation model parameter $H$ and observation noise covariance $\Sigma$. Therefore, the full set of parameters are $\phi=\{\{\theta_{i,0}\}_{i=0}^n, \{c_j\}_{j=1}^k, H, \Sigma \}$.

\paragraph{Initial Condition Parameters.}
First, we note that for the \ac{kl} divergence 
\begin{equation*}
    \KLof*{q(x \mid \tilde \theta(0))| p_0(x \mid \phi)}=-\E_{q(x \mid \tilde \theta(0)))}\left[\log \frac{ p_0(X \mid \phi)}{q(X \mid \tilde \theta(0))}\right]
\end{equation*}
in \cref{eq:app_elbo} can be computed in closed form, as both distributions are given by a product of Poisson distributions as $q(x \mid \tilde \theta(0)) = \prod_{i=1}^n \PoisDis(x_i \mid \exp(\tilde \theta_i(0)))$ and $p_0(x \mid \phi) = \prod_{i=1}^n \PoisDis(x_i \mid \exp(\theta_{i,0}))$. 
Therefore, the \ac{kl} divergence between two product Poisson distributions computes to
\begin{equation*}
    \KLof*{q(x \mid \tilde \theta(0))| p_0(x \mid \phi)}=\sum_{i=1}^n \exp(\theta_{i,0})-\exp(\tilde \theta_i(0))+ \exp(\tilde \theta_i(0))\log \frac{\exp(\tilde \theta_i(0))}{\exp(\theta_{i,0})}.
\end{equation*}
Hence, for the bound \wrt the $i$th initial condition parameter $\theta_{i,0}$ we have
\begin{equation*}
    \mathrm L(\theta_{i,0})=-\exp(\theta_{i,0}) + \exp(\tilde \theta_i(0))\theta_{i,0} +\text{const}.
\end{equation*}
Hence, computing the derivative $\frac{\partial L}{\partial \theta_{i,0}}$ and setting it to zero yields the optimal initial parameter as
\begin{equation*}
    \theta_{i,0}=\tilde \theta_i(0).
\end{equation*}

\paragraph{Rate Parameters.}
For optimizing the effective rate parameters $\{c_j\}_{j=1}^k$, we have to compute the expectation in \cref{eq:app_elbo} over the \ac{mjp} rates, which is
\begin{equation}
\label{eq:app_elbo_rates}
    \int_{0}^T \E_{q(x \mid \tilde \theta(t))}\left[ \sum_{x' \neq X} \Lambda(X,x',t) \frac{q(x' \mid \tilde \theta(t))}{q(X \mid \tilde \theta(t))} \frac{q(X\mid  \theta(t))}{q(x' \mid \theta(t))} \log \frac{\Lambda(X,x',t \mid \phi)}{\Lambda(X,x',t)} - \Lambda(X,x',t\mid \phi) \right] \diff t.
\end{equation}
First, we note that we can compute the ratios of smoothing and filtering distributions as
\begin{equation*}
     \frac{q(x' \mid \tilde \theta(t))}{q(x \mid \tilde \theta(t))} \frac{q(x \mid  \theta(t))}{q(x' \mid \theta(t))} =\prod_{i=1}^n \frac{\exp((x_i'-x_i) \tilde \theta_i(t))}{\exp((x_i'-x_i) \theta_i(t))}=\exp \left(\sum_{i=1}^n (x_i'-x_i)(\tilde \theta_i(t)- \theta_i(t)) \right).
\end{equation*}
For the rate functions $\Lambda(x,x',t)$ and $\Lambda(x,x',t \mid \phi)$ we use the equations for \iac{crn} model finding
\begin{equation*}
\begin{aligned}
    \Lambda(x,x',t)&=\sum_{j=1}^k \1(x'=x + \nu_j) c_j^{\text{old}} \prod_{i=1}^n (x_i)_{\underline{\nu}_{ij}},\\
    \Lambda(x,x',t \mid \phi)&=\sum_{j=1}^k \1(x'=x + \nu_j) c_j \prod_{i=1}^n (x_i)_{\underline{\nu}_{ij}},
\end{aligned}
\end{equation*}
where we denote by $\{c_j^{\text{old}}\}_{j=1}^k$ the set of old rate parameters, over which we do not optimize.
This yields for the expectation in \cref{eq:app_elbo_rates}
\begin{equation*}
    \begin{aligned}
    &\int_{0}^T \E_{q(x \mid \tilde \theta(t))}\left[ \sum_{x' \neq X} \Lambda(X,x',t) \frac{q(x' \mid \tilde \theta(t))}{q(X \mid \tilde \theta(t))} \frac{q(X\mid  \theta(t))}{q(x' \mid \theta(t))} \log \frac{\Lambda(X,x',t \mid \phi)}{\Lambda(X,x',t)}  -  \Lambda(X,x',t\mid \phi) \right] \diff t\\
          &\begin{multlined}
              =\int_{0}^T \E_{q(x \mid \tilde \theta(t))}\left[ \sum_{j=1}^k c_j^{\text{old}} \left\{\prod_{i=1}^n (X_i)_{\underline{\nu}_{ij}} \right\} \exp \left(\sum_{i=1}^n \nu_{ij} (\tilde \theta_i(t)- \theta_i(t)) \right)  \log  \left(\frac{c_j}{c_j^{\text{old}}}\right) \right. \\
              \left. -  c_j \prod_{i=1}^n (X_i)_{\underline{\nu}_{ij}}\right] \diff t
          \end{multlined}\\
          &\begin{multlined}
              =\sum_{j=1}^k  \int_{0}^T  \left\{   c_j^{\text{old}} \exp \left(\sum_{i=1}^n \underline{\nu}_{ij} \tilde \theta_i(t)\right) \exp \left(\sum_{i=1}^n \nu_{ij} (\tilde \theta_i(t)- \theta_i(t)) \right)  \log  \left(\frac{c_j}{c_j^{\text{old}}}\right) \right. \\
              \left. -  c_j \exp \left(\sum_{i=1}^n \underline{\nu}_{ij} \tilde \theta_i(t)\right) \right\} \diff t.
          \end{multlined}
    \end{aligned}
\end{equation*}
Hence, the bound $\mathrm L(c_j)$ \wrt the $j$th rate parameter $c_j$, can be computed as
\begin{equation*}
\begin{multlined}
      \mathrm L(c_j)=\int_{0}^T    c_j^{\text{old}} \exp \left(\sum_{i=1}^n \underline{\nu}_{ij} \tilde \theta_i(t)\right) \exp \left(\sum_{i=1}^n \nu_{ij} (\tilde \theta_i(t)- \theta_i(t)) \right)  \log  c_j \\
      -  c_j \exp \left(\sum_{i=1}^n \underline{\nu}_{ij} \tilde \theta_i(t)\right) \diff t + \text{const}.
\end{multlined}
\end{equation*}
By introducing the summary propensity statistics
\begin{equation*}
\begin{gathered}
\hat{\gamma}_j=  \frac{1}{T}\int_{0}^T c_j^{\text{old}} \exp \left(\sum_{i=1}^n \underline{\nu}_{ij} \tilde \theta_i(t)\right) c_j^{\text{old}} \exp \left(\sum_{i=1}^n \nu_{ij} (\tilde \theta_i(t)- \theta_i(t)) \right) \diff t,\\
\hat \lambda_j= \frac{1}{T}\int_{0}^T  c_j^{\text{old}} \exp \left(\sum_{i=1}^n \underline{\nu}_{ij} \tilde \theta_i(t)\right) \diff t,
\end{gathered}
\end{equation*}
we write the bound as
\begin{equation*}
     \mathrm L(c_j)= \frac{T}{ c_j^{\text{old}}} \hat \gamma_j \log c_j  - \frac{T}{ c_j^{\text{old}} }  \hat \lambda_j  c_j   + \text{const}.
\end{equation*}
Hence, computing the derivative $\frac{\partial L}{\partial c_j}$  and setting it to zero yields the optimal rate parameter as
\begin{equation*}
   c_j= \hat  \gamma_j \hat \lambda_j^{-1}.
\end{equation*}


\paragraph{Observation Model Parameters.}
The observation model parameters can be found by computing
\begin{equation*}
\begin{aligned}
&\sum_{i=1}^N \E_{q(x \mid \tilde \theta(t_i))}\left[\log p(y_i \mid X, \phi)\right] = \sum_{i=1}^N \E_{q(x \mid \tilde \theta(t_i))}\left[\log \NDis(y_i \mid H X, \Sigma)\right]\\
&\begin{multlined}
    =\sum_{i=1}^N -\frac{1}{2}  \log \vert 2 \pi \Sigma \vert 
 - \frac{1}{2} \tr\left\{ \Sigma^{-1}\left(y_i y_i^\top -  y_i \E_{q(x \mid \tilde \theta(t_i))}[ X^\top] H^\top  - H \E_{q(x \mid \tilde \theta(t_i))}[ X ] y_i^\top \right. \right.\\
 \left. \left. + H \E_{q(x \mid \tilde \theta(t_i))}[ X X^\top] H^\top \right) \right\}.
\end{multlined}
\end{aligned}
\end{equation*}
By using the product Poisson distribution $q(x \mid \tilde \theta(t_i)) = \prod_{i=1}^n \PoisDis(x_i \mid  \exp( \tilde \theta(t_i)))$, we compute the moments as $\E_{q(x \mid \tilde \theta(t_i))}[ X ]= \exp(\tilde \theta(t_i)) $ and $\E_{q(x \mid \tilde \theta(t_i))}[ X X^\top] = \diag\{\exp(\tilde \theta(t_i))\} + \exp (\tilde \theta(t_i)) \exp( \tilde \theta^\top(t_i))$. Therefore, we have the lower bound
\begin{equation*}
    \mathrm L(H, \Sigma)= - \frac{N}{2}  \log \vert 2 \pi \Sigma \vert - \frac{N}{2} \tr\left\{ \Sigma^{-1}\left(\hat{M}_{YY} - \hat{M}_{XY} H^\top  - H \hat{M}_{XY}^\top + H \hat{M}_{XX} H^\top \right) \right\}+ \text{const},
\end{equation*}
where we introduce the shorthands $\hat{M}_{YY}=\frac{1}{N} \sum_{i=1}^N y_i y_i^\top$, $ \hat{M}_{XY}= \frac{1}{N} \sum_{i=1}^N y_i \exp(\tilde  \theta^\top(t_i))$ and $\hat{M}_{XX}= \frac{1}{N} \sum_{i=1}^N  \diag\{\exp (\tilde \theta(t_i))\} + \exp(\tilde \theta(t_i)) \exp(\tilde \theta^\top(t_i))$.
Hence, computing the derivatives $\frac{\partial L}{\partial H}$ and  $\frac{\partial L}{\partial \Sigma}$, and setting them to zero yield the optimal observation model parameters as
\begin{equation*}
    \begin{aligned}
    &H=\hat{M}_{XY} \hat{M}_{XX}^{-1}, &&\Sigma=\hat{M}_{YY}-  \hat{M}_{XY} H^\top - H \hat{M}_{XY}^\top   + H \hat{M}_{XX} H^\top.
    \end{aligned}
\end{equation*}
Note that, this is very similar to the case of a linear Gaussian state space model, for more see \cite{sarkka2013bayesian}.
\section{ADDITIONAL EXPERIMENTAL RESULTS}
\label{sec:app_experiments}
In this section, we provide further details on the experiments presented in the main document, along with additional experiments. 

\subsection{Benchmark Tasks}

\subsubsection{Lotka-Volterra Model}
For the Lotka-Volterra task discussed in the experiments section of the main paper, we evaluate our method and the baseline methods on $100$ sample trajectories with time horizon $T=300$. Each latent trajectory gets observed at $10$ randomly chosen, non-equidistant time points with linear Gaussian measurements of the latent state, \ie, we assume the observation likelihood $p(y_i \mid x) = \NDis(y_i \mid H x, \Sigma)$. The learning rate for the expectation propragation algorithm, also called dampening parameter is set to $\epsilon = 0.05$ for this and all other experiments. 
We summarize the parameters in \cref{table_params_lv}. The results of the experiments are highlighted in the main paper.

\begin{table}[ht]
    \centering
    \caption{Parameters of the Lotka-Volterra experiments}
    \begin{tabular}{|c|c|}
        \hline
        \textbf{Parameter} & \textbf{Value} \\
        \hline
        rate $c_1$  & 0.005 \\
        rate $c_2$  & 0.001 \\
        rate $c_3$  & 0.005 \\
        observation matrix $H$  & $\begin{bmatrix}
1 & 0\\
0& 1 
\end{bmatrix}$ \\
        observation covariance matrix $\Sigma$ & $\begin{bmatrix}
1 & 0\\
0& 1 \end{bmatrix}$ \\
        \hline
    \end{tabular}
    \label{table_params_lv}
\end{table}
\subsubsection{Motility Model}
\begin{figure}
\centering \includegraphics{figures/motility_with_particle/lgend_aistats_EP_vs_SMC.pdf}\\
\includegraphics[width=.33\textwidth]{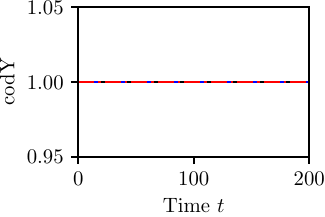}\hfill
\includegraphics[width=.33\textwidth]{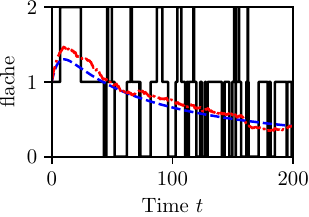}\hfill
\includegraphics[width=.33\textwidth]{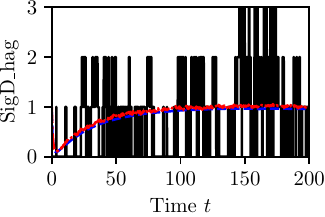}\\
\includegraphics[width=.33\textwidth]{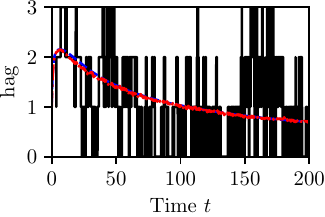}\hfill
\includegraphics[width=.33\textwidth]{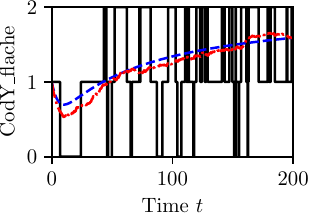}\hfill
\includegraphics[width=.33\textwidth]{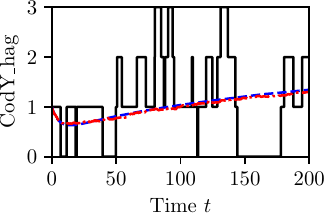}
    \caption{Additional results of the motility model. Here, the posterior mean
tracks roughly the time-average of the ground-truth signal. However, since for this species the effective realizations are very low, random fluctuations play a more significant role, making accurate tracking more challenging. The results of our method closely aligns with the \ac{smc} results. 
}
    \label{fig:wilk2}
\end{figure}

For our motility model experiment, we consider a sample trajectory with time horizon $T=200$. The latent trajectory is observed at $5$ randomly chosen, non-equidistant time points with linear Gaussian measurements of the SigD species.
The parameters of the experiment are summarized in \cref{table_params_motility}. The additional results of the motility model experiment are depicted in \cref{fig:wilk2}.
This experiment highlights the particle degeneracy issue of the \ac{smc} method, which we describe shortly. In our implementation of the \ac{smc} method, we start with $N_s=10000$ uniformly weighted samples. At each observation point, we update the weights based on the observation likelihood and then resample the trajectories according to the updated weights. This process leads to repeated resampling of the early section of the trajectories, resulting in a significant reduction in the number of distinct trajectories in these sections.
In our experiment, the section until the first observation time point of the latent estimate is described by only $1734$ distinct trajectories out of the $10000$ samples. While this still provides a good approximation of the ground truth, the number of distinct trajectories decreases even further for longer or more complex tasks, necessitating more samples to maintain accuracy.
Although \ac{smc} with a large number of samples serves as a reliable approximation of the ground truth for comparison in our study, the number of samples required increases significantly with model complexity and time horizon, making \ac{smc} less scalable and less practical for higher-dimensional or longer-horizon problems. In contrast, the computational cost of our method scales linearly with the time horizon, as it primarily depends on solving a system of \acp{ode}, where the number of required computations increases proportionally with the length of the time horizon.

\begin{table}[ht]
    \centering
    \caption{Parameters of the motility experiment}
    \begin{tabular}{|c|c|}
        \hline
        \textbf{Parameter} & \textbf{Value} \\
        \hline
        rate $c_1$  & 0.1 \\
        rate $c_2$  & 0.0002 \\
        rate $c_3$  & 1. \\
        rate $c_4$  & 0.0002 \\
        rate $c_5$  & 1.0 \\
        rate $c_6$  & 0.0002 \\
        rate $c_7$  & 0.01 \\
        rate $c_8$  &  0.1\\
        rate $c_9$  & 0.02 \\
        rate $c_{10}$  & 0.1 \\
        rate $c_{11}$  & 0.01 \\
        rate $c_{12}$  & 0.1 \\
        observation covariance  $\sigma^2$ & 100 \\
        \hline
    \end{tabular}
            \label{table_params_motility}
\end{table}

\subsubsection{Gene Transcription and Translation}
As an additional benchmark we consider a stochastic model representing the gene transcription and translation \citep{anderson2015stochastic}. Transcription refers to the process of copying information encoded in the DNA to a messenger RNA (mRNA). The translation of an mRNA by a ribosome yields proteins.

The model is defined by the following reactions:
\begin{equation*}
\begin{aligned}
 &\text{G} \xrightarrow{c_1} \text{G} + \text{M},
 &&\text{M} \xrightarrow{c_2} \text{M} + \text{P},
  &&&\text{M} \xrightarrow{c_3} \emptyset,
   &&&&\text{P} \xrightarrow{c_4} \emptyset,
\end{aligned}
\end{equation*}
where G represents gene, M the mRNA and P the proteins. The reactions represent the transcription, translation, degradation of mRNA and the degradation of proteins, respectively.

Similar to the Lotka-Volterra experiment, we infer the approximate posterior for $100$ sample trajectories with time horizon $T=8$.
Each latent trajectory is observed at $10$ randomly chosen, non-equidistant time points with linear Gaussian measurements of the protein species. We summarize the parameters in \cref{table_params_gene}. We again compare the results to the baseline results by \begin{inlineitemize}
\item a single \ac{ffbs} iteration using the entropic matching method without \ac{ep},
\item a Gaussian \ac{ads} based on the chemical Langevin equation, similar to the method described by \citet{cseke2016expectation},
\item the moment-based \ac{vi} method proposed by \citet{wildner2019moment}, and
\item an exact smoothing algorithm based on a truncation of the system, which serves as ground truth for our comparison.
\end{inlineitemize}

\begin{table}[ht]
    \centering
    \caption{Parameters of the gene transcription and translation experiment}
    \begin{tabular}{|c|c|}
        \hline
        \textbf{Parameter} & \textbf{Value} \\
        \hline
        rate $c_1$  & 200 \\
        rate $c_2$  & 10 \\
        rate $c_3$  & 25 \\
        rate $c_4$  & 1\\
        observation covariance  $\sigma^2$ & 10 \\
        \hline
    \end{tabular}
    \label{table_params_gene}
\end{table}

\begin{figure}
    \centering
    \includegraphics[width=0.5\linewidth]{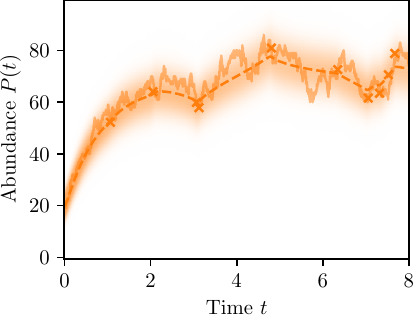}
    \caption{Simulation of the Gene Expression model. The protein species is visualized in orange. Dashed lines denote variational posterior mean, solid lines denote ground truth trajectory, the background indicates the inferred marginal state probabilities and the crosses indicate observations.}
    \label{fig:gene}
\end{figure}

\begin{table}[h!]
\centering
\caption{Mean squared error in posterior mean averaged over trajectories and time for the gene transcription and translation experiment}
\begin{tabular}{|c|c|c|c|}
\hline
\ac{ep} (Ours) &\ac{ffbs} Entropic &G \ac{ads} &MB\ac{vi}\\ 
\hline
\textbf{0.1919} &2.9912 & 652.68& 2.2687\\ 
\hline
\end{tabular}
\label{table_results_gene}
\end{table}
\cref{table_results_gene} shows the mean squared error in the posterior mean of the approximate methods compared to the exact posterior mean of the truncated system.
Similar to the Lotka-Volterra experiment, our method demonstrates superior performance overall. Notably, the Gaussian \ac{ads} performs poorly, which we attribute to its inappropriate modeling choice of representing low population counts with Gaussian distributions.
We visualize the result of one sample trajectory in \cref{fig:gene}.

\subsubsection{Enzyme Kinetics}
Finally, as last benchmark we study the enzyme kinetics model, a standard model in which an enzyme catalyzes the conversion of some substrate to  product \citep{anderson2015stochastic}.

The model is defined by the following reactions:

\begin{equation*}
\begin{aligned}
&\text{S} + \text{E} \xrightarrow{c_1} \text{SE},
&&\text{SE} \xrightarrow{c_2} \text{S} + \text{E},
&&&\text{SE} \xrightarrow{c_3} \text{P} + \text{E},
\end{aligned}
\end{equation*}
where S is the substrate species, E the enzyme, SE an enzyme-substrate and P the product species.

Again, we infer the approximate posterior for $100$ sample trajectories with time horizon $T=20$.
Each latent trajectory is observed at $10$ randomly chosen, non-equidistant time points with linear Gaussian measurements of the protein and substrate species. We summarize the parameters in \cref{table_params_enzyme}. We again compare the results to the previously mentioned baselines.

\begin{table}[ht]
    \centering
    \caption{Parameters of the enzyme kinetics experiment}
    \begin{tabular}{|c|c|}
        \hline
        \textbf{Parameter} & \textbf{Value} \\
        \hline
        rate $c_1$  & 0.05 \\
        rate $c_2$  & 0.5 \\
        rate $c_3$  & 0.5 \\
        observation covariance matrix $\Sigma$ & $\begin{bmatrix}
10 & 0\\
0& 10 \end{bmatrix}$ \\
        \hline
    \end{tabular}
    \label{table_params_enzyme}
\end{table}

\begin{figure}
    \centering
    \includegraphics[width=0.5\linewidth]{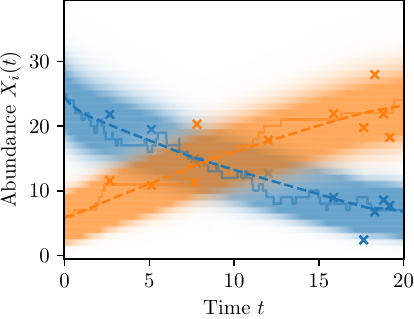}
    \caption{ Simulation of the Enzyme Kinetics model. The abundance of the substrate species is visualized in blue, the product species in orange. Dashed lines denote variational posterior mean, solid lines denote ground truth trajectory, the background indicates the inferred marginal state probabilities and the crosses indicate observations.}
    \label{fig:enzyme}
\end{figure}

\begin{table}[h!]
\centering
\caption{Mean squared error in posterior mean averaged over trajectories and time for the gene transcription and translation experiment}
\begin{tabular}{|c|c|c|c|}
\hline
\ac{ep} (Ours) &\ac{ffbs} Entropic &G \ac{ads} &MB\ac{vi}\\ 
\hline
\textbf{0.3339} &1.3432 & 7.0121& 0.6091\\ 
\hline
\end{tabular}
\label{table_results_enzyme}
\end{table}
\cref{table_results_enzyme} shows the mean squared error in the posterior mean of the approximate methods compared to the exact posterior mean of the truncated system. Like in the previous cases, out method demonstrates superior performance in estimating the posterior mean.
We visualize the result of one sample trajectory in \cref{fig:enzyme}.

\subsection{Parameter Learning}
In this section we analyze the task of inferring the latent state when not all parameters of the model are known. To address this, we employ the previously described approximate \ac{em} algorithm to jointly infer both the parameters and the latent state.
We focus on experiments with unknown rate parameters. Closed form solutions for the M-Step of the \ac{em} algorithm are derived in \cref{app:param_learning_crn}.
Notably, in our experiments we find that using the entropic matching method with a single \ac{ffbs} iteration yields parameter estimation results comparable to those using \ac{ep} method with entropic matching. This enables a significantly faster algorithm, as only one \ac{ffbs} iteration needs to be computed per \ac{em} step. Finally for the latent state estimation we use the proposed \ac{ep} algorithm with the resulting parameters of the final \ac{em} step. 

We apply the \ac{em} algorithm to the Lotka-Volterra model, the Gene transcription and translation task and the enzyme kinetics model, using the same data as in the previous experiments. Specifically, for each task, we run the \ac{em} algorithm on each of the 100 sample trajectories with the same observations as described earlier.
For the Lotka-Volterra model, we assume all rate parameters to be unknown; for the gene translation and transcription task we assume $c_2$ to be unknown; and for the enzyme kinetics task we treat $c_3$ as unknown.

We initialize all parameters with starting guesses in the same order of magnitude as their true values, ensuring a realistic but non-trivial starting point for the inference process. \cref{table_params_em} shows the true parameter values, the initial guesses for the \ac{em} algorithm and the estimated values, averaged across the 100 sample trajectories.
We observe that our proposed method yields reasonably accurate results for the Lotka-Volterra task, where all rate parameters are unknown and very good results for the other tasks where only one parameter is unknown.
\begin{table}[ht]
    \centering
    \caption{Estimated Parameter Values}
    \begin{tabular}{|c|c|c|c|}
        \hline
        \textbf{Parameter} & \textbf{True Value} & \textbf{Initial Value} & \textbf{Estimated Value}\\
        \hline
        LV  $c_1$  & 0.005 & 0.002 & 0.0077 \\
        LV $c_2$  & 0.001 & 0.002 &0.0012 \\
        LV $c_3$  & 0.005 & 0.002& 0.0064 \\
        \hline
        Gene $c_2$  & 10.0 & 5.0 & 9.963\\
        \hline
        Enzyme $c_3$ & 0.5 & 0.1 & 0.520\\
        \hline
    \end{tabular}
    \label{table_params_em}
\end{table}

Further very interesting are the results for the latent state inference. We compute the approximate posterior mean for all tasks and compare it to the results of the exact method based on truncation, assuming full knowledge of the parameters. The error averaged over time and trajectories is summarized in \cref{table_state_em}. When comparing these results with \cref{table_results_enzyme,table_results_gene,table_lv_results}, we find that our proposed method for  joint inference of parameters and latent state demonstrates superior performance in estimating the mean compared to the baselines methods that utilize full knowledge of the parameters. Naturally, the \ac{ep} algorithm with full knowledge of the parameters performs even better. 
This underscores the efficacy of the \ac{em} algorithm.

\begin{table}
\centering
\caption{Mean squared error in posterior mean averaged over trajectories and time for all tasks}
\begin{tabular}{|c|c|c|}
\hline
LV & Gene &Enzyme\\ 
\hline
0.7955 &1.0936 & 0.4289\\ 
\hline
\end{tabular}
\label{table_state_em}
\end{table}

\subsection{Computational Cost}
The computational cost of our proposed method is primarily driven by the repeated solution of the \acp{ode} in \cref{eq:filter_crn,eq:smoother_crn}. Since we derived closed-form solutions for the right-hand sides of these equations, we can evaluate these \acp{ode} efficiently and fast.
For the numerical integration, we utilized the Runge-Kutta-Fehlberg method from the SciPy package. All computations were performed on an Apple M1 chip.

Similarly, the computational cost of the moment-based \ac{vi} method by \citet{wildner2019moment} is dominated by repeatedly solving \acp{ode} for a forward and a backward pass. However, this method approximates the first and second-order moments of the latent state distribution, which increases the dimensionality of the \ac{ode} system significantly, making this approach less scalable for high-dimensional problems.

The Gaussian \ac{ads} approximates first and second-order moments similar to the \ac{vi} method, and therefore encounters the same scalability issues as the number of species increases. However the Gaussian \ac{ads} is significantly faster, as it requires only one forward and backward pass.

In contrast, the computational cost of the \ac{smc} method depends on the number of samples and the cost of generating a sample trajectory. 
As discussed earlier, the particle degeneracy issue can significantly increase the number of particles required to maintain an accurate approximation, making this approach less scalable, particularly for longer time horizons or high-dimensional problems.

\subsection{Discussion}
The experimental results demonstrate the effectiveness and scalability of our proposed method for latent state inference and parameter learning in complex models. However, it is important to acknowledge that other methods have their own merits, and the choice of approach depends largely on the specific use case and model characteristics.

For latent state inference in \acp{mjp} with possibly unknown parameters and low to moderate population sizes, we recommend our proposed method.
In contrast, for systems with large population sizes where the underlying \ac{mjp} can be well-approximated by \iac{sde}, the Gaussian \ac{ads} offers a suitable alternative.

When the focus extends beyond posterior marginals to the full posterior path, the moment-based variational inference method proposed by \citet{wildner2019moment} is a strong option. Alternatively, the mean-field approach proposed by \citet{opper2007variational} can be employed, which however, is limited to models where only one species changes per jump. In systems where multiple species change simultaneously (as in our LV model), this assumption breaks down, leading to issues with absolute continuity, as discussed by \citet{wildner2019moment}. The neural variational inference method by \citet{seifner2023neural} either integrates the chemical master equation, which is only tractable for small systems or they use the mean-field approach by \citet{opper2007variational}.

\ac{smc} methods \citep{doucet2001sequential}, while computationally intensive and subject to particle degeneracy over long time horizons, can still provide reliable results when configured with a sufficiently large number of particles. They remain a useful choice, particularly when high accuracy is required for state estimation over shorter sequences.
Finally, if parameter learning is the primary goal, without the need for detailed latent state inference, \ac{mcmc} methods provide a Bayesian approach that offers posterior distributions over parameters instead of point estimates \citep{golightly2011bayesian,golightly2015delayed,lowe2023accelerating}.

\end{document}